\documentclass{ieeeaccess}
\usepackage{amsmath,amssymb,amsfonts}
\usepackage{algorithmic}
\usepackage{graphicx}
\usepackage{textcomp}

\usepackage{hyperref}
\newcommand{\orc}{\includegraphics[height=\fontcharht\font`A]{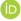}}
\usepackage[
    sorting=none,
    bibstyle=ieee
]{biblatex}
\DeclareMathOperator*{\argmax}{argmax}
\usepackage{subcaption}
\usepackage{tabularray}
\usepackage[export]{adjustbox}
\usepackage{multicol}
\usepackage{multirow}
\usepackage[normalem]{ulem}
\newtheorem{prop}{Proposition}
\usepackage{comment}
\usepackage[en-GB]{datetime2}
\DTMlangsetup[en-GB]{ord=omit}
\DTMsettimestyle{iso}

\usepackage{bm}
\makeatletter
\AtBeginDocument{\DeclareMathVersion{bold}
\SetSymbolFont{operators}{bold}{T1}{times}{b}{n}
\SetSymbolFont{NewLetters}{bold}{T1}{times}{b}{it}
\SetMathAlphabet{\mathrm}{bold}{T1}{times}{b}{n}
\SetMathAlphabet{\mathit}{bold}{T1}{times}{b}{it}
\SetMathAlphabet{\mathbf}{bold}{T1}{times}{b}{n}
\SetMathAlphabet{\mathtt}{bold}{OT1}{pcr}{b}{n}
\SetSymbolFont{symbols}{bold}{OMS}{cmsy}{b}{n}
\renewcommand\boldmath{\@nomath\boldmath\mathversion{bold}}}
\makeatother

\def\BibTeX{{\rm B\kern-.05em{\sc i\kern-.025em b}\kern-.08em
    T\kern-.1667em\lower.7ex\hbox{E}\kern-.125emX}}

\begin{document}
\history{Received 13 December 2024, accepted 5 January 2025, date of publication 8 January 2025. Author's version, excludes proofing from the Journal.}
\doi{10.1109/ACCESS.2025.3527227}

\title{Feature-Centered First Order Structure Tensor Scale-Space in 2D and 3D}
\author{\uppercase{Pawel Tomasz Pieta}\href{https://orcid.org/0009-0005-7634-6627}{\orc}, \uppercase{Anders Bjorholm Dahl}\href{https://orcid.org/0000-0002-0068-8170}{\orc}, (Member, IEEE), \uppercase{Jeppe Revall Frisvad}\href{https://orcid.org/0000-0002-0603-3669}{\orc}, \uppercase{Siavash Arjomand Bigdeli}\href{https://orcid.org/0000-0003-2569-6473}{\orc}, \uppercase{Anders Nymark Christensen}\href{https://orcid.org/0000-0002-3668-3128}{\orc}}

\address{Department of Applied Mathematics and Computer Science, Technical University of Denmark, 
2800 Kongens Lyngby, Denmark}

\tfootnote{This work was supported by Innovation Fund Denmark, project 0223-00041B (ExCheQuER).}

\markboth
{P. T. Pieta \headeretal: Feature-Centered First Order Structure Tensor Scale-Space in 2D and 3D}
{P. T. Pieta \headeretal: Feature-Centered First Order Structure Tensor Scale-Space in 2D and 3D}

\corresp{Corresponding author: Pawel Tomasz Pieta (e-mail: papi@dtu.dk).}

\begin{abstract}
The structure tensor method is often used for 2D and 3D analysis of imaged structures, but its results are in many cases very dependent on the user's choice of method parameters. We simplify this parameter choice in first order structure tensor scale-space by directly connecting the width of the derivative filter to the size of image features. By introducing a ring-filter step, we substitute the Gaussian integration/smoothing with a method that more accurately shifts the derivative filter response from feature edges to their center. We further demonstrate how extracted structural measures can be used to correct known inaccuracies in the scale map, resulting in a reliable representation of the feature sizes both in 2D and 3D. Compared to the traditional first order structure tensor, or previous structure tensor scale-space approaches, our solution is much more accurate and can serve as an out-of-the-box method for extracting a wide range of structural parameters with minimal user input.
\end{abstract}


\titlepgskip=-21pt

\maketitle

\section{Introduction}
\IEEEPARstart{S}{tructure} tensor is a well-established image processing method used widely for structural analysis of images. It can often be applied directly to the raw image data, providing rich and explainable descriptions of the analyzed structure. Its simplicity and versatility make it a popular choice across a diverse range of fields~\cite{LucasJ,Frangi1998,Sato1997,Faraklioti2005}, with recent applications focusing on 3D image analysis~\cite{straumit2016a, Jeppesen2020, larsson2014a, wang2019a, roesler2024a}.

Importantly, structure tensor continues to be an important, and often irreplaceable tool, despite the recent advances in deep learning-based methods. Since structural analysis is typically not predictive (without a specified target, e.g. segmentation or classification) \cite{Jeppesen2020, larsson2014a}, the potential training labels would have to directly reflect the structural properties that are practically impossible to label by hand, especially in real, complex materials. With the lack of ground-truth data, the alternative would be to train a model that directly emulates the structure tensor output. Fundamentally, structure tensor consists of simple convolution operations, so an attempt to apply deep learning would likely be less efficient and provide the same or degraded results, with any potential improvements only on the edges of different materials, where structure tensor has some difficulties.

In its most recognized form, structure tensor calculation starts with selecting two parameters that approximately correspond to the size of investigated structures. This choice can often be challenging if the prior information about the imaged object is limited. The problem becomes even more pronounced when the structures span a wide range of sizes, often resulting in noisy or inaccurate representations.

Being Gaussian-based, the size limitations of structure tensor can be addressed using the scale-space theory~\cite{Lindeberg1994, weickert1999a}. However, existing solutions are often complex and rely on assumptions derived from edge detection methods. We propose redefining the structure tensor and its scale-space from the perspective of structural features, which are not isolated edges but rather paired edges forming objects (e.g., fibers). With the adoption of this feature-oriented approach, we revise the theoretical framework of the method, introducing modifications to the filtering steps at a single scale and refining the scale normalization and selection process. Our experiments demonstrate that the resulting method is more robust and consistent, returns scale-invariant results, and is easier to effectively use.

\section{Related work} \label{sec:related_work}

The structure tensor was first introduced independently by Big{\"u}n and G{\"o}sta~\cite{Bigun1987} and Förstner and G{\"u}lch~\cite{ForstnerST}. While neither explicitly used the term "structure tensor," their theoretical formulations are closely aligned with it. Further foundations were laid by Knutsson~\cite{knutsson1989a}, with a very similar formulation widely adopted for the purposes of corner and edge detection~\cite{Harris1988, Lindeberg1996}. Early applications of the structure tensor in images include orientation analysis~\cite{LucasJ}, line and vessel segmentation~\cite{Frangi1998,Sato1997}, or seismic data analysis~\cite{Faraklioti2005}.

The rise in computational power and the growing availability of volumetric data facilitated the extension of the structure tensor to 3D images~\cite{Kovalev1999, Wang2004}. This trend is largely continued, with recent applications focusing on the analysis of large and complex 3D volumes in disciplines such as material science~\cite{straumit2016a, nguyen2018a, Jeppesen2020} or medicine~\cite{larsson2014a, Reichardt2021, wang2019a,roesler2024a}. Advances in computational techniques have also enabled the acceleration and parallelization of its computation, making the structure tensor method more accessible and useful~\cite{Jeppesen2021}.

The scale-space theory was most comprehensively developed by Lindeberg~\cite{Lindeberg1994}, though its origins remain a topic of debate~\cite{weickert1999a}. Over the years, it has become a recognized and established method, contributing to advancements in a wide range of algorithms and fields~\cite{alvarez1999a,florack2000a,lindeberg2011a,worrall2019,lindeberg2022a}. Within the context of the structure tensor scale-space, the most mathematically robust definition is again provided by Lindeberg~\cite{Lindeberg1994}, involving a complex two-step scale exploration process. Other, more case-specific formulations have been proposed~\cite{Wang2004,zhou2012a,xu2016a}, but these often interpret the concept of scale less strictly, diverging from the formal constraints of scale-space theory.

\section{Problem formulation} \label{sec:problem}

Depending on the chosen source and the task at hand, we can define a set of different structure tensor formulations, primarily based on various derivative filters. The most widely-adapted form uses two Gaussian filters: a first order Gaussian derivative for extracting a smooth gradient representation and a smoothing Gaussian filter for integrating the neighbourhood information. As a result, a typical calculation starts with a choice of the sizes of the two filters, based on the size of features in the image (investigated structural elements, e.g. fibers, or vessels). The gradient information is then collected for each pixel in the form of a square matrix representing the structure factor, from which we can extract the local orientation or the shape of the investigated features.

\subsection{Scale-Space}  \label{sec:mod_scale-space}
\IEEEpubidadjcol
In many applications, performing a structure tensor calculation using a single set of parameters (i.e.,\ at a single scale) is sufficient. Oftentimes, the image features have a known, relatively consistent size, which means that the method can provide a relatively accurate result for these simple cases. This behaviour is caused by a combination of two factors: the use of smoothing and a practical limitation of a first order Gaussian derivative filter -- due to having only one zero-crossing, it does not align exactly to any given feature size (compared e.g.\ to the Laplacian filter). 

When a filter is not aligned well with a feature, the directional response is distorted to a degree that is hard to quantify. The seeming insensitivity of the algorithm to varying feature sizes often makes it hard to choose the optimal scale for a given image. When searching the parameter space, we often do not notice a strong change in the overall result, whereas it is still statistically noticeable when the results are used in downstream tasks. Such a dilemma is especially true in cases where the features have variable or unknown sizes. All these problems mean that a single scale result is not stable and trustworthy enough to be confidently used for more complex cases. We need a technique that extracts a response for a feature found optimally by some stable metric.

An obvious solution is to use a scale-space approach~\cite{Lindeberg1994} where the algorithm is computed at multiple scales, and a maximum response is chosen for a given interest point among the scales. In the case of a Gaussian-based scale-space, the differently scaled responses need to be normalized in order to accurately compare them. As stated in \cite[p.~4]{lindeberg1998a}, \emph{"the amplitudes of spatial derivatives in general decrease with scale"}, which for the Gaussian-based scale-space representation $L$ of a signal can be corrected with:
\begin{equation}
\label{eq:lind_norm}
L_{\xi^m}(\cdot,\sigma) = \sigma^{m \gamma}L_{x^m}(\cdot,\sigma)
\end{equation}
where $\sigma$ is the Gaussian standard deviation, i.e., the scale, $L(\cdot,\sigma)$ is the scale-normalized representation, $x$ is a partial derivative in some dimension, $\xi$ is a scale-normalized version of this derivative, $m$ is the order of the derivative (equal to 1 in our case), and $\gamma$ is a heuristic scaling parameter.

In the idealized case, $\gamma=1$, which was used in the first and most influential work on the scale-space structure tensor~\cite[p.~359]{Lindeberg1994}. In this work, Lindeberg used a complex two-step approach, explained by the need to define two separate filter sizes. First, the filter sizes are set to have a fixed ratio, and an optimal scale is chosen using measures related to derivative magnitude, typically the trace of the square structure matrix. Next, an optimal size of the Gaussian derivative filter is found according to some criterion (e.g.\ maximum anisotropy). Importantly, Lindeberg highlighted that the second, integrating Gaussian is the main source of the feature size (scale correlation). This approach was however only used on a predefined set of interest points, not for all pixels (in a dense approach).

The value of $\gamma$ in connection with the 1st order Gaussian derivatives was later questioned~\cite[p.~10]{Lindeberg1996}. A value of $\frac{1}{2}$ was shown to be optimal by fitting the filter to a diffuse Gaussian step edge. This value has recently been used successfully as a part of a dense scale selection algorithm~\cite{Lindeberg2018}. One of the main contributions of our work is to challenge this approach.

\subsection{Proposed approach}

\begin{figure}[!t]
     \centering
     \begin{subfigure}[b]{0.29\columnwidth}
         \centering
         \includegraphics[width=\textwidth]{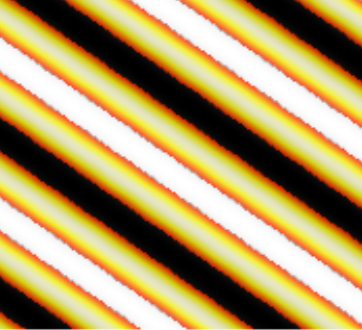}
         \caption{Features in edge detection.}
         \label{fig:featureDefinition_a}
     \end{subfigure}
     \begin{subfigure}[b]{0.345\columnwidth}
         \centering
         \includegraphics[width=\textwidth]{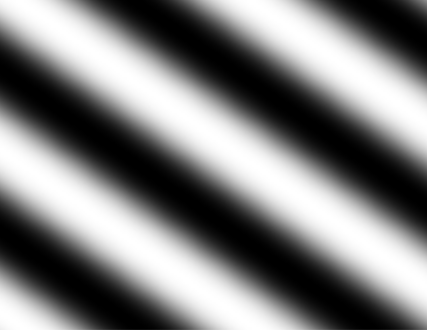}
         \caption{Sample image. \\\hspace{\textwidth}}
         \label{fig:featureDefinition_b}
     \end{subfigure}
     \begin{subfigure}[b]{0.29\columnwidth}
         \includegraphics[width=\textwidth]{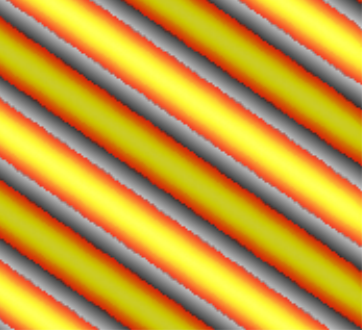}
         \caption{Features in structural analysis.}
         \label{fig:featureDefinition_c}
     \end{subfigure}
    \caption{Features and their approximate width on a simplified fiber image, defined for edge detection and structural analysis. In edge detection, the feature is usually defined as a gradient of the fiber's edge. In structural analysis, the feature is the fiber itself.}
    \label{fig:featureDefinition}
\end{figure}

When considering the size of a structural feature, one would more naturally think of e.g.\ the width of a fiber than the gradient of the fiber's edge (Fig.~\ref{fig:featureDefinition}). Because of this, we claim that instead of fitting the filter to a diffuse Gaussian step edge, it is more descriptive to fit it to a rectangular function. Our results show that in such a case we must have $\gamma > 1$ to achieve a linear relation between feature and filter size. 

With this paradigm change, we can move the source of feature size (scale correlation) from the integrating Gaussian to the Gaussian derivative filter. This change is important, as there is an inherent flaw in the use of a Gaussian filter in the second smoothing step. Its size is typically set to be significantly bigger than the derivative filter, to account for the fact that the strongest derivative response is placed on the edges of the feature. Despite the Gaussian size and due to its shape, the majority of information will still be collected from the feature's center, which can be weak, noisy or incorrect. To address this potential inaccuracy, we propose to use a ring-like filter that is aligned with an expected feature size, collecting primarily the edge information. The original Gaussian filter can still be used afterwards simply as a smoothing step.

Finally, we acknowledge the problem of shifted scale response for blob-like features, described by Lindeberg~\cite[p.~12]{Lindeberg1996}, and observe that the structure tensor shape measures can be used to adjust the scale map and provide a dense structure tensor scale-space representation.

\section{Theory} \label{sec:int_theory}

The structure tensor formulation, in its modern form, is usually derived from a second order Taylor expansion of an $N$ dimensional image $I$ in the neighbourhood of a point $x$ using a scale of $s$~\cite{Frangi1998} 
\begin{equation}
\label{eq:taylor}
I(x +\delta x,s) \approx I(x,s) + \delta x^T \nabla_{s} + \delta x^T H_{s} \delta x\,,
\end{equation}
where $\delta x$ is a small offset into the neighbourhood. The resulting matrices: the second-moment matrix $\nabla_{s}$ and the Hessian matrix $H_{s}$ are representations of the first and second order structure tensors, respectively, in the point $x$. These matrices are in practice computed by convolving the image with corresponding Gaussian derivative filters at a given scale in order to suppress noise influence~\cite{Sato1997}. An additional Gaussian smoothing step often follows to integrate and average the neighbourhood information~\cite{LucasJ,Harris1988}. Such an approach results in two parameters of the method, which in the traditional approach are
\begin{itemize}
    \item $\sigma$ - standard deviation of the derivative Gaussian. This defines the degree of noise suppression.
    \item $\rho$ - standard deviation of the integrating Gaussian. For the first order structure tensor typically $\rho\geq2\sigma$, which results in combining the response from edges of the structural element to a ``mean'' direction of the whole element.
\end{itemize}

An eigendecomposition of the matrices enables the extraction of a set of ordered eigenvalues $(\lambda_1 \leq \lambda_2 \leq ... \leq \lambda_N)$ with corresponding eigenvectors, where the smallest eigenvalue marks the direction of least change. In the following analysis, the eigenvalues are often used to create measures that describe the shape of detected features \cite{STintro}. In 2D, it is possible to calculate the anisotropy $A$ of the detected features by
\begin{equation}
\label{eq:anis_2D}
A = 1- \frac{\lambda_1}{\lambda_2} \,.
\end{equation}
In 3D, measures are commonly borrowed from diffusion tensor theory. One example is fractional anisotropy:
\begin{equation}
\label{eq:anis_3D}
\mathrm{FA} = \sqrt{\frac{3}{2}\left(\frac{(\lambda_1-\hat{\lambda})^2+(\lambda_2-\hat{\lambda})^2+(\lambda_3-\hat{\lambda})^2}{\lambda_1^2+\lambda_2^2+\lambda_3^2}\right)} \,,
\end{equation}
as a representation of anisotropy, where $\hat{\lambda} = (\lambda_1 + \lambda_2 + \lambda_3)/3$. Another example is a set of three measures: linearity $m_\ell$, sphericity $m_s$ and planarity $m_p$~\cite{Westin2002,STintro}, where
\begin{equation}
\label{eq:meas_3D}
m_\ell=\frac{\lambda_2-\lambda_1}{\lambda_3}, m_p = \frac{\lambda_3-\lambda_2}{\lambda_3}, m_s = \frac{\lambda_1}{\lambda_3}\,.
\end{equation}
Due to the properties of Gaussian derivative filters, the first order structure tensor is the most widely used for structure and orientation analysis. Its second order version is not capable of providing precise directional data but can be more accurate in segmenting anisotropic features~\cite{Frangi1998}. In the following sections, the structure tensor algorithm always refers to its first order version.

\section{Method}

\subsection{Scale-Space}

\begin{figure}[!t]
\centering
\includegraphics[width=\columnwidth]{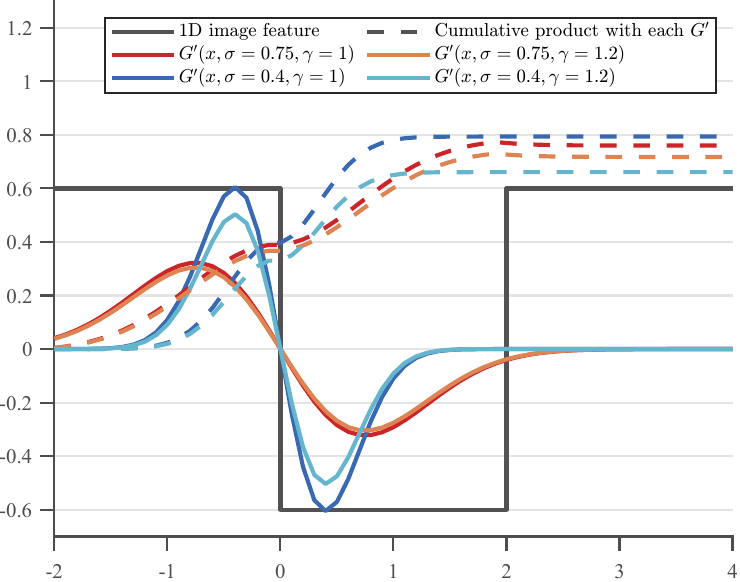}
\caption{Gaussian derivative filters aligned with a feature edge. With default normalization, filters with $\sigma\rightarrow 0$ return a higher convolution response (red and blue). Setting $\gamma>1$ yields a higher value for $\sigma > 0$ (orange and cyan).}
\label{fig:1D_scale-space_ex}
\end{figure}

To ensure a correlation between the size of the derivative filter and the feature width, we consider a simplified model of a 1D feature in the shape of a rectangular function with width $x_f$. This can be interpreted as a cross-section of a noise-free image of e.g.\ a fiber. Given that the 1st order Gaussian derivative has only one zero-crossing, it is clear that a normalized response with $\gamma=1$ will be maximized for $\sigma \rightarrow 0$ (Fig.~\ref{fig:1D_scale-space_ex}). To change this, we artificially increase the response from wider filters by using $\gamma>1$. 

Starting with an explicit definition of a Gaussian $g$ and its normalized 1st order derivative $g'_{\mathrm{norm}}$, we derive the convolution response $P$ for the edge of the suggested feature $\Pi(\frac{x-\frac{x_f}{2}}{x_f})$, where $\Pi$ is the rectangular function, and $-\frac{x_f}{2}$ shifts the function's rising edge to zero:
\begin{equation}
\label{eq:gauss}
g(x,\sigma) = \frac{1}{\sigma\sqrt{2\pi}}e^{\frac{-x^2}{2\sigma^2}}, \quad
g'_{\mathrm{norm}}(x,\sigma, \gamma) = \frac{\partial g(x,\sigma)}{\partial x}\sigma^\gamma,
\end{equation}
\begin{eqnarray}
\lefteqn{P(x_f,\sigma,\gamma) = \int_{-\infty}^{0} g'_{\mathrm{norm}}(x,\sigma,\gamma) \,dx} \nonumber \\
 &&- \int _{0}^{x_f}g'_{\mathrm{norm}}(x,\sigma,\gamma) \,dx\ + \int _{x_f}^{\infty}g'_{\mathrm{norm}}(x,\sigma,\gamma) \,dx\,.\quad \label{eq:rect_conv}
\end{eqnarray}

Solving for $\sigma^{\ast} = \argmax_{\sigma}P$, we find a filter-to-feature size ratio $t$ that is static for a given $\gamma$. Using that
\begin{align} 
\gamma = \frac{1}{t^2 (e^{\frac{1}{2t^2}} -1)} + 1 \,, \quad t=\frac{\sigma^{\ast}}{x_f} \label{eq:gamma_ratio}\\
\mbox{\textit{Derivation in Appendix \ref{App:proof_eq_norm_1}}}\notag
\end{align}
we find
\begin{align}
t = \frac{1}{\sqrt{1-\gamma-2W_{-1}(\frac{1-\gamma}{2}e^{\frac{1-\gamma}{2}})}} \,, \label{eq:t_gamma_ratio}\\
\mbox{\textit{Derivation in Appendix \ref{App:proof_eq_norm_2}}}\notag
\end{align}
where $W_{-1}(\cdot) \leq -1$ is a branch of the Lambert $W$ function. For $(\gamma \rightarrow 1 \land \gamma > 1)$, we have $W_{-1}(\cdot) \rightarrow -\infty$, and consequently $t \rightarrow 0$, which forces $\sigma^{\ast} \rightarrow 0$, thus confirming the assumption demonstrated on Fig.~\ref{fig:1D_scale-space_ex}.

In practice, useful $\gamma$ values are between 1 and 1.5. The higher the value of $\gamma$, the bigger the part of the optimal filter outside the feature. Intuitively, the size of that section should be small in order to limit the amount of outside information being picked up. On the other hand, $\gamma$ cannot be too small, as it would result in a very weak ability to align with the edge of the feature in the presence of noise. In the following sections, we use $\gamma=1.2$ as a middle ground between the two edge cases, resulting in the following filter-to-feature size relation:
\begin{equation}
\label{eq:gamma_practical_ratio}
\sigma^{\ast} \approx 0.372\, x_f \,.
\end{equation}

\subsection{Ring Filter} \label{sec:ring_filter}

\begin{figure}[!t]
     \centering
     \begin{subfigure}[b]{0.22\textwidth}
         \centering
         \includegraphics[width=\textwidth]{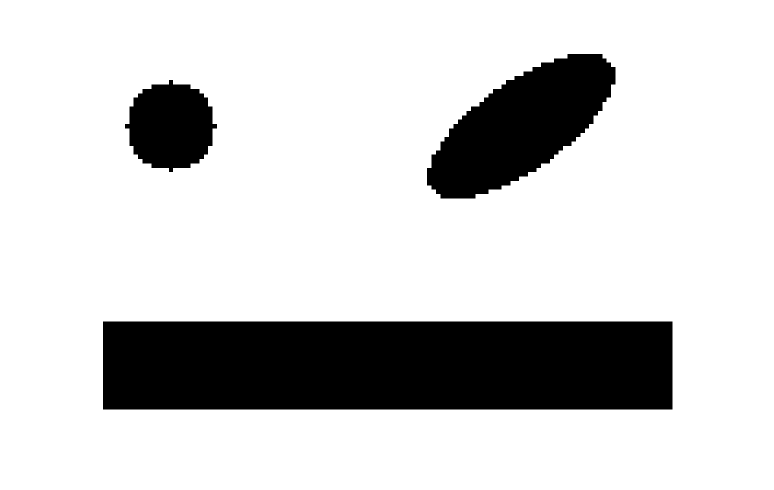}
         \caption{Sample image.}\quad
         \label{fig:ex1_sample_image}
     \end{subfigure}
     \begin{subfigure}[b]{0.22\textwidth}
         \centering
         \includegraphics[width=\textwidth]{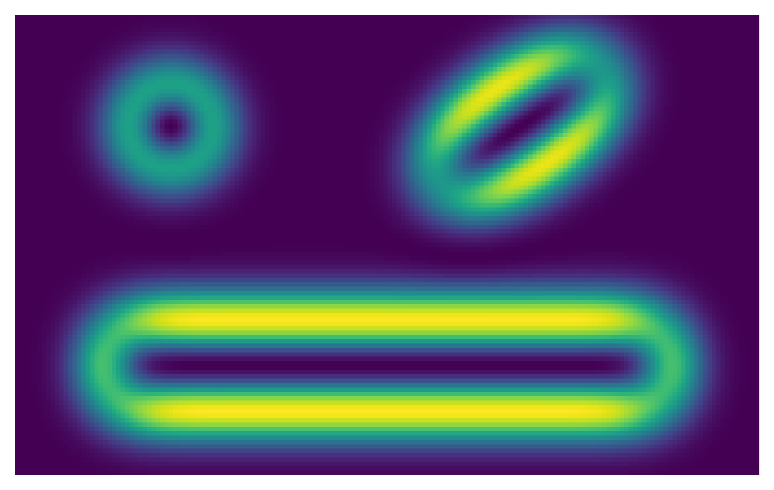}
         \caption{$\mathrm{tr}(\nabla_{0,s})$ after the derivative Gaussian.}
         \label{fig:ex2_derivative_response}
     \end{subfigure}
     \begin{subfigure}[b]{0.22\textwidth}
         \includegraphics[width=\textwidth]{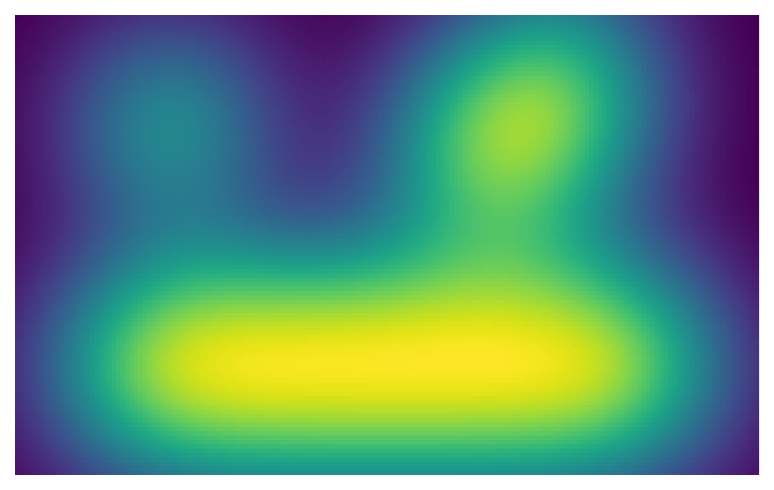}
         \caption{$\mathrm{tr}(\nabla_{s})$ with Gaussian smoothing, $\rho=2\sigma$.}
         \label{fig:ex3_smoothing_orig}
     \end{subfigure}
     \begin{subfigure}[b]{0.22\textwidth}
         \centering
         \includegraphics[width=\textwidth]{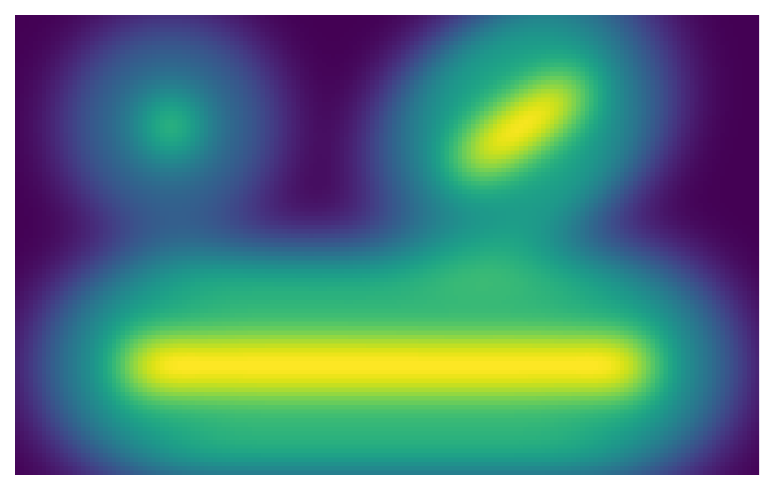}
         \caption{$\mathrm{tr}(\nabla_{s})$ after applying a ring filter, no smoothing.}
         \label{fig:ex4_ring_filter}
     \end{subfigure}
    \caption{Effect of running the structure tensor algorithm on a sample image with a circular (isotropic), a linear (anisotropic), and an elliptical feature, all with the same width (20px), using a single appropriate scale. The raw derivative filter response is placed on the edges of the features. Both strong smoothing and a ring filter shift the response to the object center, but the ring filter provides stronger peaks on object centers and a lower smoothing effect. The isotropic feature gets a lower response in all cases.}
    \label{fig:ex_image_ring_effect}
\end{figure}

In the conventional structure tensor method, strong smoothing is employed primarily to shift the filter response from a feature's edges to its center (Fig.~\ref{fig:ex2_derivative_response},\ref{fig:ex3_smoothing_orig}). We instead suggest the use of a ring filter (Fig.~\ref{fig:ex4_ring_filter}), which would align with the edges of a feature and avoid distortion from the noisy information in the center of a feature. The ring filter size has to be adjusted to the expected feature size and the ring width should be relatively small. The most basic design would consist of a simple binary ring filter, but this would fail in conjunction with the scale-space approach, as it has been shown that only Gaussian-based filtering operations are capable of generating a scale-space~\cite[p.~14]{Lindeberg1994}.

To satisfy the requirement of a Gaussian, we construct a ring filter by using the difference of two differently-sized unnormalized Gaussian filters: 
\begin{equation}
\label{eq:ring_filter}
 R(x,\sigma_R,k) = e^{\frac{-x^2}{2\sigma_R^2}} - e^{\frac{-x^2}{2(\sigma_R k)^2}}, \quad k \in (0,1).
\end{equation}
This approach enables us to have a zero value in the center of the filter (as opposed to the Laplacian of a Gaussian, for example). Additionally, the separability of the Gaussian kernel~\cite[p.~114]{Burger2009} allows for a much faster calculation than if using a pre-defined binary filter.

\begin{figure}[!t]
\centering
\includegraphics[width=3.6in]{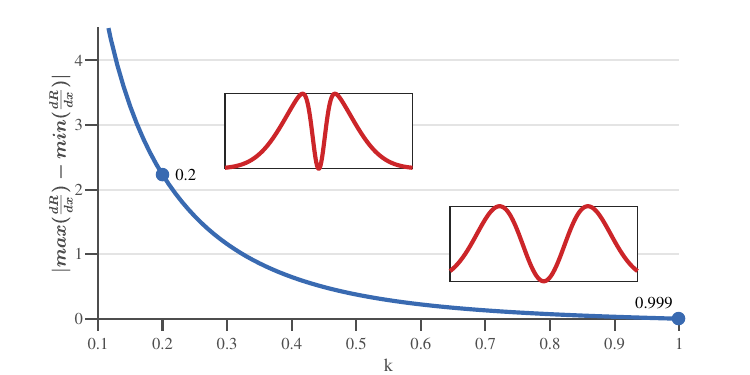}
\caption{Difference between the slope angles of the outer and inner sides of the ring filter, depending on the $k$ parameter, with visualization of 1D ring filters for two values of $k$. The slopes of the filter are more uniform for $k$ approaching $1$.}
\label{fig:ringSlope}
\end{figure}

The parameter $k$ is the size ratio between the two filters. In practice, we use $k = 0.999$ because the closer $k$ is to $1$, the smaller the difference in the slope between the outer and the inner side of the ring (see Fig.~\ref{fig:ringSlope}). A change in the standard deviation $\sigma_R$ affects the width of the filter:
\begin{align}
\argmax_x R(x,\sigma_R,k) = \pm \sigma_R k \sqrt{\frac{2 \ln(k^2)}{k^2-1}} \label{eq:ring_filter_max} \,.\\
\mbox{\textit{Derivation in Appendix \ref{App:proof_eq_ring}}}\notag
\end{align}
We link this relation to the $\sigma^{\ast}$ value of the $g'_{\mathrm{norm}}$ filter by
\begin{equation}
\label{eq:ring_filter_max_t}
\argmax_x R(x,\sigma_R,0.999) \approx \pm 1.414 \,\sigma_R \,,
\end{equation}
\begin{equation}
\label{eq:ring_filter_to_sigma}
\sigma^{\ast} = t x_f = t \big[ 2 \argmax_x R(x,\sigma_R,0.999) \big] \approx 1.052 \,\sigma_R \,,
\end{equation}
\begin{equation}
\label{eq:ring_filter_to_sigma_2}
\sigma_R \approx 0.95\,\sigma^{\ast} \,.
\end{equation}

\subsection{Scale Map Correction} \label{sec:mod_scale_corr}

A side effect of using a first order derivative filter is that the response from an isotropic (blob-like) feature is weaker than from a corresponding (in size and intensity) anisotropic (fiber-like) feature (Fig. \ref{fig:ex_image_ring_effect}). This effect is known to cause differences in the scale selection for the two feature types, where the optimal scale for the isotropic one is significantly lower~\cite[~p. 12]{Lindeberg1996}. The magnitude of this effect depends on the curvature of the edge of the analyzed object. We argue that ratios of the structure tensor eigenvalues are useful for mitigating this problem. More specifically, we use the anisotropy measure $A$ introduced in Sec.~\ref{sec:int_theory} (Eq.~\ref{eq:anis_2D}). 

Lindeberg~\cite{Lindeberg1996} demonstrates that the difference between the detected scales depends on the chosen scale-space discriminator, and derives an equation of this effect for an equivalent of the structure tensor matrix trace:
\begin{equation}
\label{eq:lind_iso_scale}
\sigma_{\mathrm{iso}}^{\ast} = \frac{\gamma}{3-\gamma} \sigma_{\mathrm{anis}}^{\ast}\underset{\gamma=1.2}= \frac{2}{3} \, \sigma_{\mathrm{anis}}^{\ast} \,,
\end{equation}
where $\sigma_{\mathrm{iso}}^{\ast}$ and $\sigma_{\mathrm{anis}}^{\ast}$ are the optimal scales for isotropic and anisotropic features of the same size. We choose to use this discriminator, as it is a very direct representation of the gradient magnitude. We consider the optimal scale of an arbitrary feature a linear blend of $\sigma^{\ast}_{\mathrm{iso}}$ and $\sigma^{\ast}_{\mathrm{anis}}$ and use $A$ as the blending factor:
\begin{equation}
\label{eq:blending_iso}
\sigma^{\ast} = (1 - A) \sigma^{\ast}_{\mathrm{iso}} + A \sigma^{\ast}_{\mathrm{anis}} \,.
\end{equation}
A version of Lindeberg's relation generalized for any given feature is then
\begin{equation}
\label{eq:iso_scale_ratio_gen}
\frac{\sigma^{\ast}}{\sigma^{\ast}_{\mathrm{anis}}} = (1 - A)\frac{\sigma^{\ast}_{\mathrm{iso}}}{\sigma^{\ast}_{\mathrm{anis}}} + A
\end{equation}
or, using Eq.~\ref{eq:lind_iso_scale} for $\gamma = 1.2$, we have
\begin{equation}
\label{eq:iso_scale_corr_gen}
\sigma_{\mathrm{anis}}^{\ast} 
= \frac{\sigma^{\ast}}{(1-A)\frac{2}{3} + A}
= \frac{\sigma^{\ast}}{1-\frac{1}{3}(1-A)} \,.
\end{equation}

A separate side effect comes from the ring filter and how it aligns with an anisotropic feature. The ring diameter that returns the strongest response for a set of two parallel lines is intuitively bigger than the feature width. The exact ratio $t_{\mathrm{anis}} = \sigma^{\ast}/x_{f,\mathrm{anis}}$ between the optimal scale and the anisotropic feature width depends on a number of factors, such as the strength of the derivative response, ring filter and derivative response widths and how they scale with increasing radius/feature width. 
To avoid the complications coming from the Gaussian function, we do not find an exact formula for $t_{\mathrm{anis}}$. Instead, we approximate the widths of the ring filter and $\mathrm{tr}(\nabla_{0,s})$ after using the Gaussian derivative filter and convert them to uniform kernels. Under this assumption, we prove that $t_{\mathrm{anis}}$ is independent of the feature size (Appendix~\ref{App:proof_of_prop_1}, Proposition~\ref{prop:ring-line_ratio}). 

\begin{figure}[!t]
\centering
\includegraphics[width=3.6in]{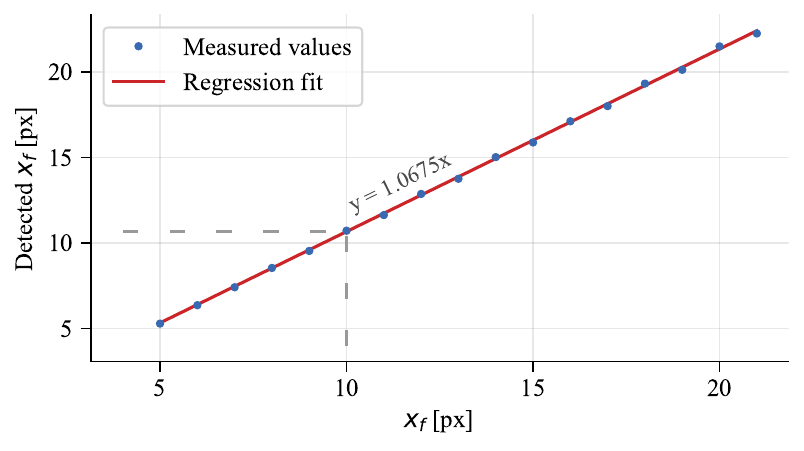}
\caption{Numerical quantification of the relation between the width of an anisotropic feature (rectangular binary line) and detected feature width (at the center of it), caused by the alignment with the ring filter. The relation is independent of the feature width. The small deviations can most likely be attributed to numerical errors caused by representations of even and odd-sized features.}
\label{fig:line-ring_ratio}
\end{figure}

Based on empirical results (Fig.~\ref{fig:line-ring_ratio}), we find that, when using the complete filter definitions, the following correction ratio is needed
\begin{equation}
\label{eq:anis_scale_corr}
\frac{\sigma_{\mathrm{anis}}^{\ast}}{\sigma_{\mathrm{anis,corr}}^{\ast}} \approx 1.0675 \,.
\end{equation}
As in Eqs.~\ref{eq:blending_iso}-\ref{eq:iso_scale_corr_gen}, we use $A$ as a blending factor to generalize the correction:
\begin{equation}
\label{eq:blending_anis_1}
\sigma^{\ast} = (1 - A) \sigma_{\mathrm{anis,corr}}^{\ast} + A \sigma_{\mathrm{anis}}^{\ast} \,,
\end{equation}
\begin{equation}
\label{eq:blending_anis_2}
\frac{\sigma^{\ast}}{\sigma_{\mathrm{anis,corr}}} = (1 - A)  + A \frac{\sigma_{\mathrm{anis}}^{\ast}}{\sigma_{\mathrm{anis,corr}}^{\ast}} \,,
\end{equation}
\begin{equation}
\label{eq:anis_scale_corr_gen}
\sigma_{\mathrm{anis,corr}} \approx \frac{\sigma^{\ast}}{(1 - A)  + 1.0675 A} = \frac{\sigma^{\ast}}{1 + 0.0675 A}\,.
\end{equation}

\begin{figure}[!t]
     \centering
     \begin{subfigure}[b]{0.23\textwidth}
         \centering
         \includegraphics[width=\textwidth]{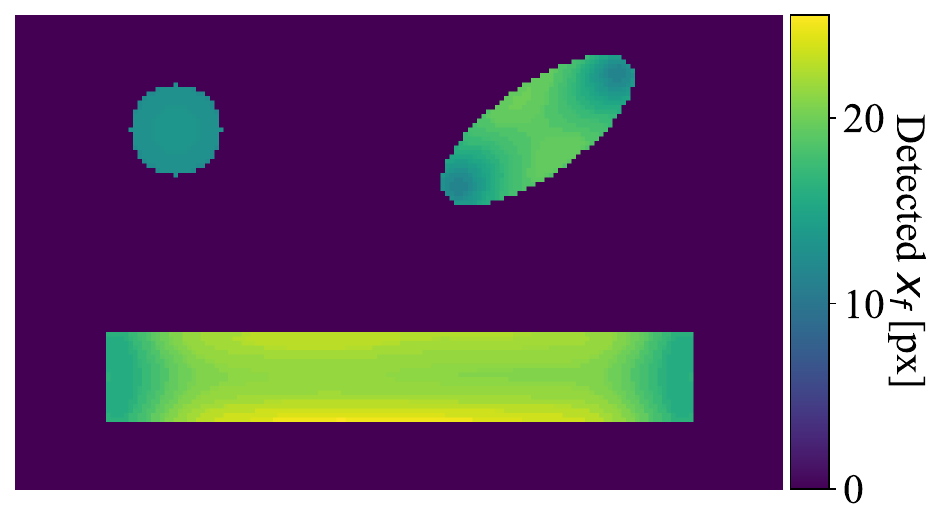}
         \caption{Raw scale image.}
         \label{fig:scaleEx_noNorm_im}
     \end{subfigure}
     \begin{subfigure}[b]{0.23\textwidth}
         \centering
         \includegraphics[width=\textwidth]{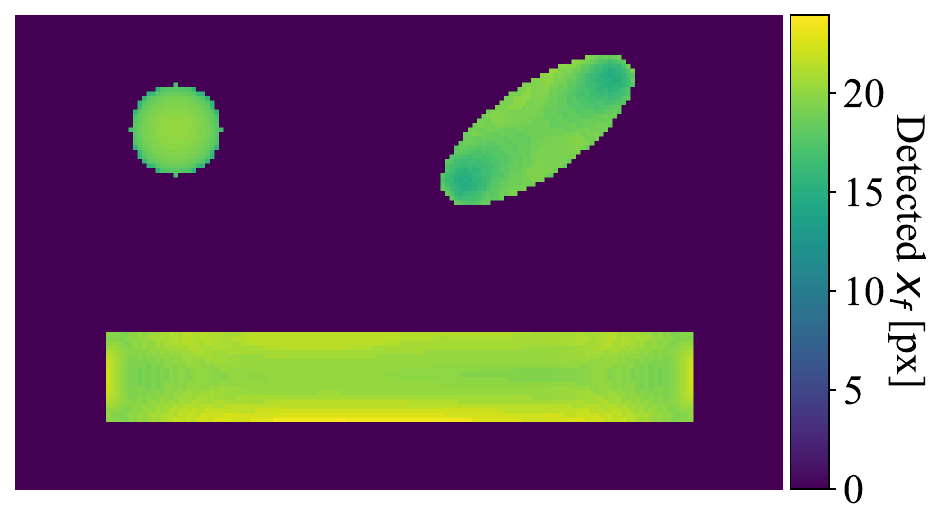}
         \caption{Scale image after correction.}
         \label{fig:scaleEx_norm_im}
     \end{subfigure}
     \begin{subfigure}[b]{0.49\textwidth}
         \includegraphics[width=\textwidth,trim={0 3.5mm 0 0},clip]{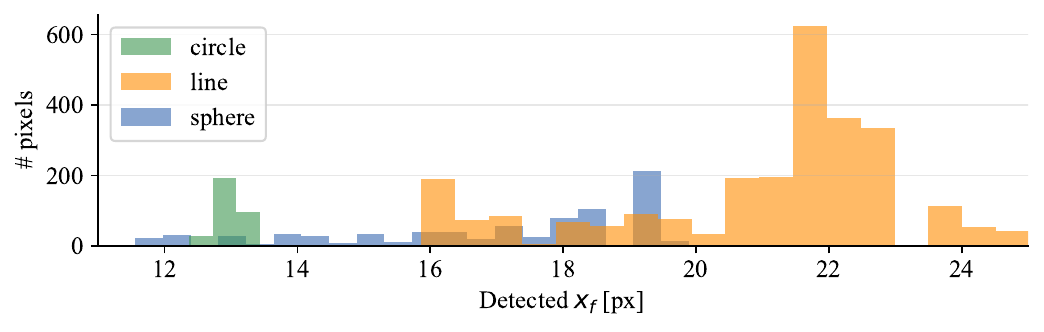}
         \caption{Scale histogram without correction.}
         \label{fig:scaleEx_noNorm_hist}
     \end{subfigure}
     \begin{subfigure}[b]{0.49\textwidth}
         \centering
         \includegraphics[width=\textwidth,trim={0 3.5mm 0 0},clip]{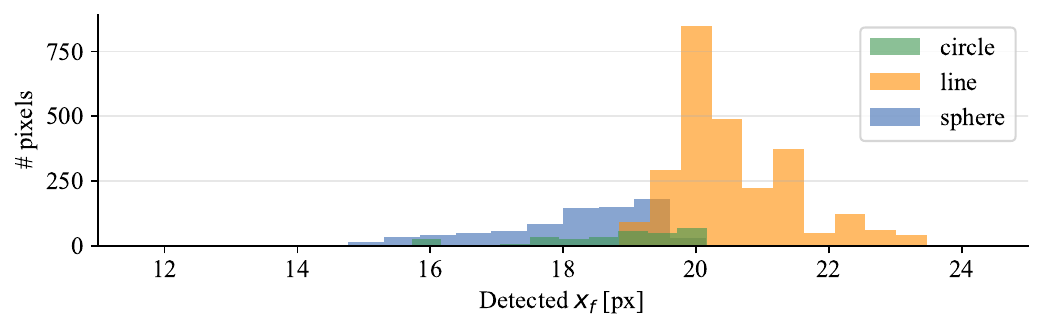}
         \caption{Scale histogram after correction.}
         \label{fig:scaleEx_norm_hist}
     \end{subfigure}
    \caption{Effects of the scale map correction for the sample image in Fig.~\ref{fig:ex1_sample_image} (excluding areas outside the features). Before correction (a and c), both the histogram and the scale image return smaller scale values for isotropic features. After the correction (b and d), this result is significantly mitigated, especially for the centers of the features.}
    \label{fig:scaleEx}
\end{figure}

The two correction equations (Eqs.~\ref{eq:iso_scale_corr_gen},\ref{eq:anis_scale_corr_gen}) are caused by two independent effects and can be combined. Supposing $S_{\mathrm{2D}}$ is a dense scale map estimate, we can then correct detected scales using:
\begin{equation}
\label{eq:scale_corr_2d}
S_{\mathrm{2D,corr}}  \approx \frac{S_{\mathrm{2D}}}{(1+0.0675 A)(1-\frac{1}{3}(1-A))} \,.
\end{equation}
The effects of this correction are demonstrated in Fig.~\ref{fig:scaleEx}. It is important to note that the most accurate scale estimation can only be found approximately along the skeleton of the feature. Moreover, the values found there correspond to the local distance to edges, which is demonstrated well for the elliptical feature. Moving away from the skeleton and closer to the edges, scale values increase for the isotropic features and decrease for anisotropic features. This again is related to how the ring filter aligns locally with the derivative response.

Scale estimation in 3D is much more complicated than its 2D version. Instead of two primary shapes (isotropic circles and anisotropic lines), we have three shapes: spherical, linear, and planar. Importantly, the width of the planar features is the distance between its two planes, whereas, for linear features, the width is approximately the radius of a cylinder. As a result, the scale detection inaccuracy does not correlate linearly with the anisotropy score and it cannot be effectively used for scale correction (Table~\ref{tab:3D_shapes}). On the other hand, seeing how the degree of scale correction is correlated to the sphericity, linearity, and planarity of the features, it is only natural to correct the scale map using the metrics for these shapes defined in Sec.~\ref{sec:int_theory}. 

Due to the higher complexity of the 3D case, we treat the choice of correction parameters as an optimization problem. The purpose of our optimization is to minimize the difference between the scale values in the centers of the three primary feature types (all assumed to have the same width). We tested the method on a 2D case and obtained correction parameters similar to those in Eq.~\ref{eq:scale_corr_2d}. We then adjusted the equation to include the shape metrics (Eqs.~\ref{eq:meas_3D}), and the optimization resulted in the following 3D scale correction: 
\begin{align}
\label{eq:scale_corr_3d}
& S_{\mathrm{3D,corr}}  \notag \\  & \approx \frac{S_{\mathrm{3D}}}{0.53\,(1 +0.0158\,m_s)(1 + m_p)(1 + 0.327\,m_\ell)}.
\end{align}

\begin{table}
\begin{center}
\caption{Basic 3D structural shapes, their anisotropy and detected scale, assuming they are the same size. Compared to the 2D case, the anisotropy is not linearly correlated to the detected scale distortion.}
\label{tab:3D_shapes}
\begin{tblr}{
  cells={valign=m,halign=c},
  row{1}={font=\bfseries},
  vline{2} = {-}{0.08em},
  hline{2} = {-}{0.16em},
}
Feature type & Feature sketch & FA & Detected scale
\\
spherical & \includegraphics[width=0.05\textwidth,valign=c]{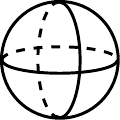} & $\downarrow$ & $\downarrow$ \\ 
planar & \includegraphics[width=0.05\textwidth,valign=c]{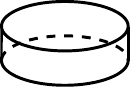} & $-$ & $\uparrow$ \\ 
linear & \includegraphics[width=0.05\textwidth,valign=c]{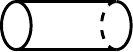} & $\uparrow$ & $-$ \\ 
\end{tblr}
\end{center}
\end{table}

\section{Experiments}

The first focus of our experiments is to demonstrate the advantages of our method as compared with the conventional scale-space approach~\cite[p.~359]{Lindeberg1994}. We chose the conventional approach as a baseline, as it seems the most advanced and commonly recognized implementation of the method. For the comparison, we use the simple artificial image in Fig.~\ref{fig:ex1_sample_image} and more complex 2D and 3D datasets.

The advantage of our method is most easily explored in cases with variable-sized features and features of unknown size. In addition, we test the behaviour of our method when run on an incorrect scale range and suggest solutions for correcting the range.

\subsection{Artificial Data}

\begin{figure}[!t]
    \begin{tblr}{
      cells={valign=m,halign=c},
      colspec={QQQQ},
    }
    \vspace{0.10cm}  & Anisotropy & Orientation  \\
    \rotatebox[origin=c]{90}{$\sigma=6.4$} \rotatebox[origin=c]{90}{$\rho=2\sigma$} & \includegraphics[width=0.21\textwidth,valign=c]{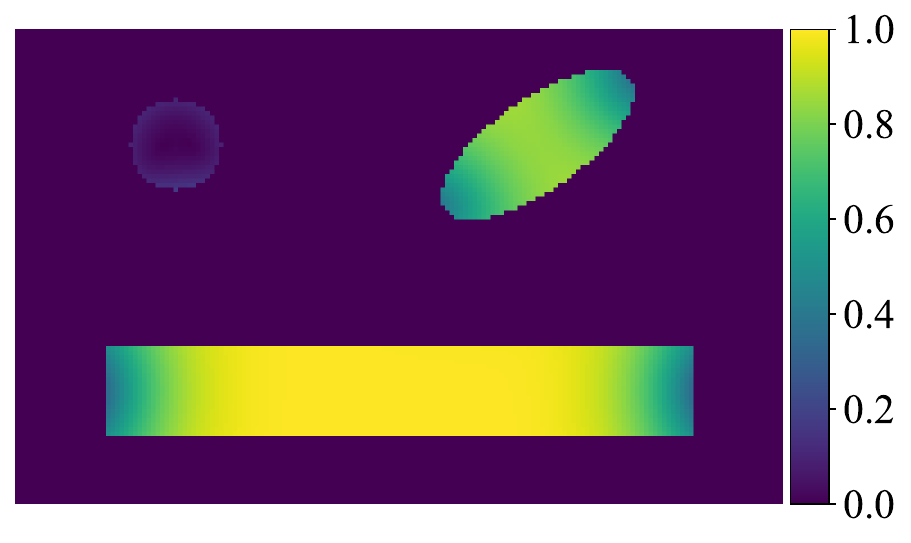} & \includegraphics[width=0.18\textwidth,valign=c]{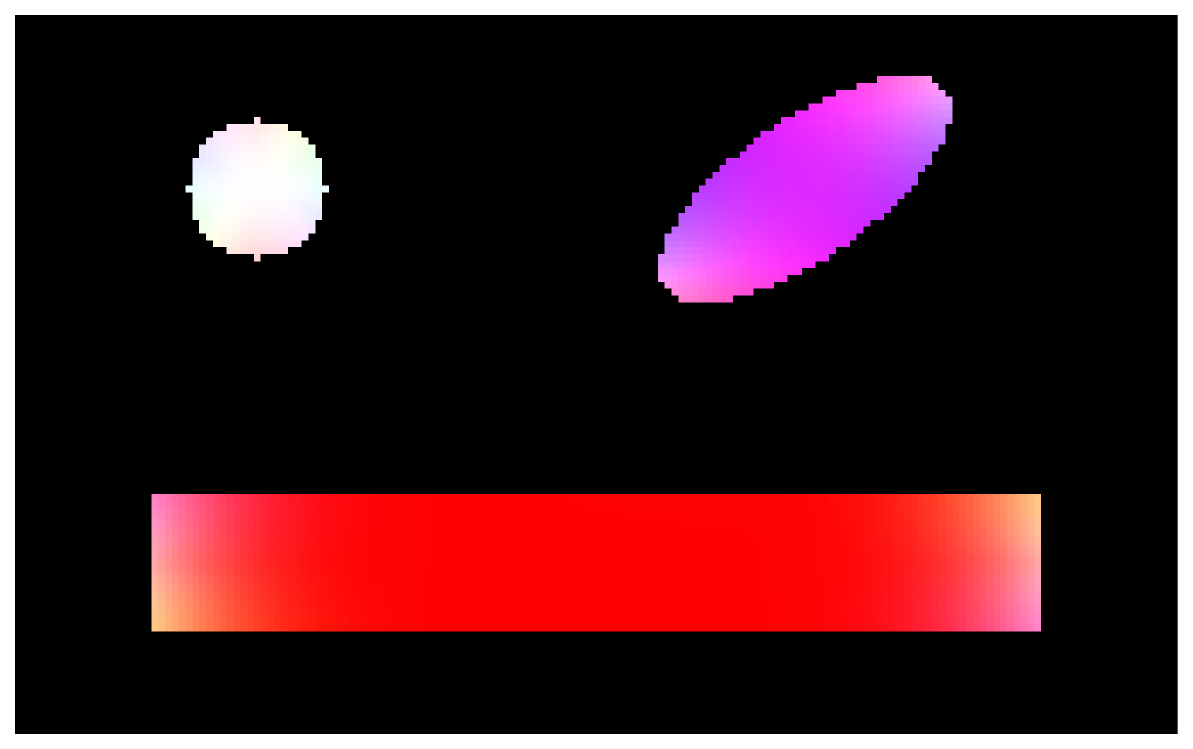}\\
    \rotatebox[origin=c]{90}{Baseline} \rotatebox[origin=c]{90}{scale-space} & \includegraphics[width=0.21\textwidth,valign=c]{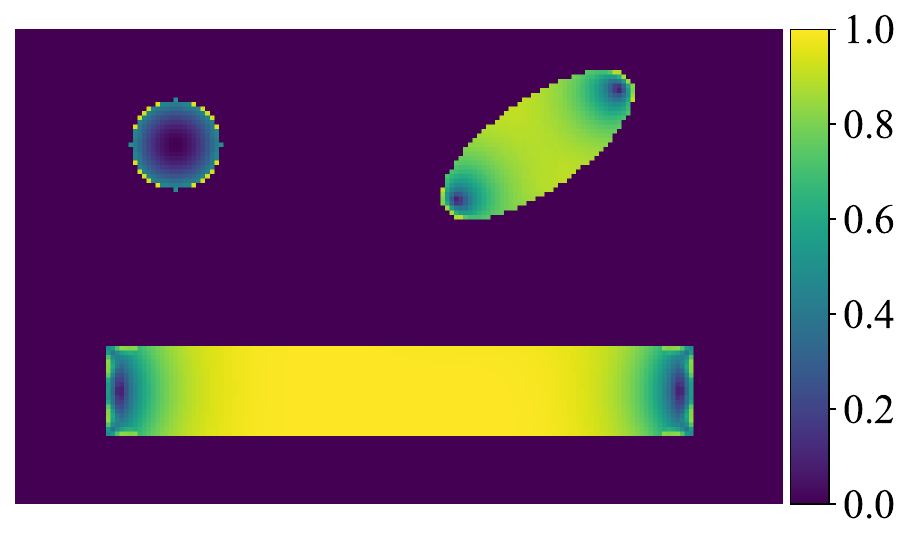} & \includegraphics[width=0.18\textwidth,valign=c]{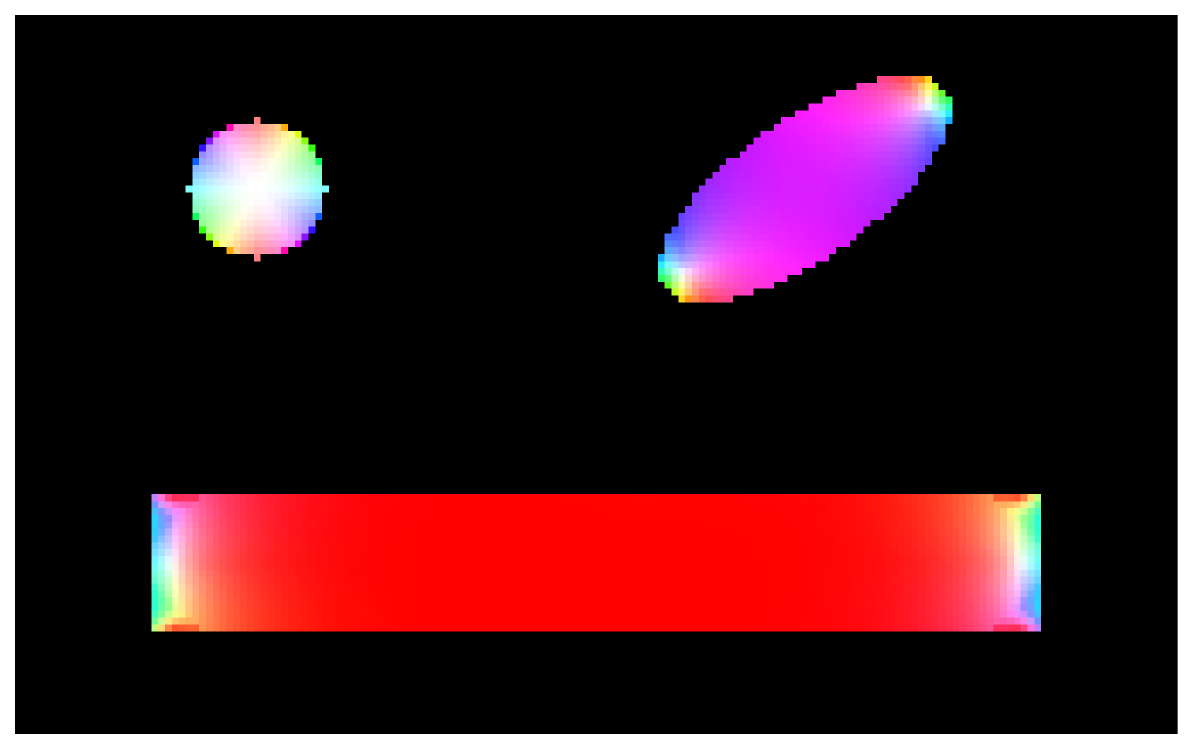} \\
    \rotatebox[origin=c]{90}{Proposed} \rotatebox[origin=c]{90}{scale-space} & \includegraphics[width=0.21\textwidth,valign=c]{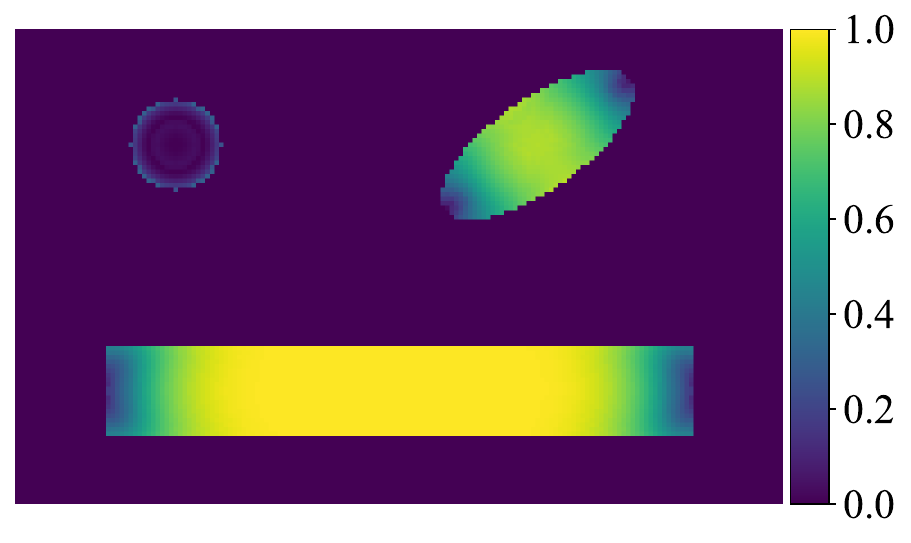} & \includegraphics[width=0.18\textwidth,valign=c]{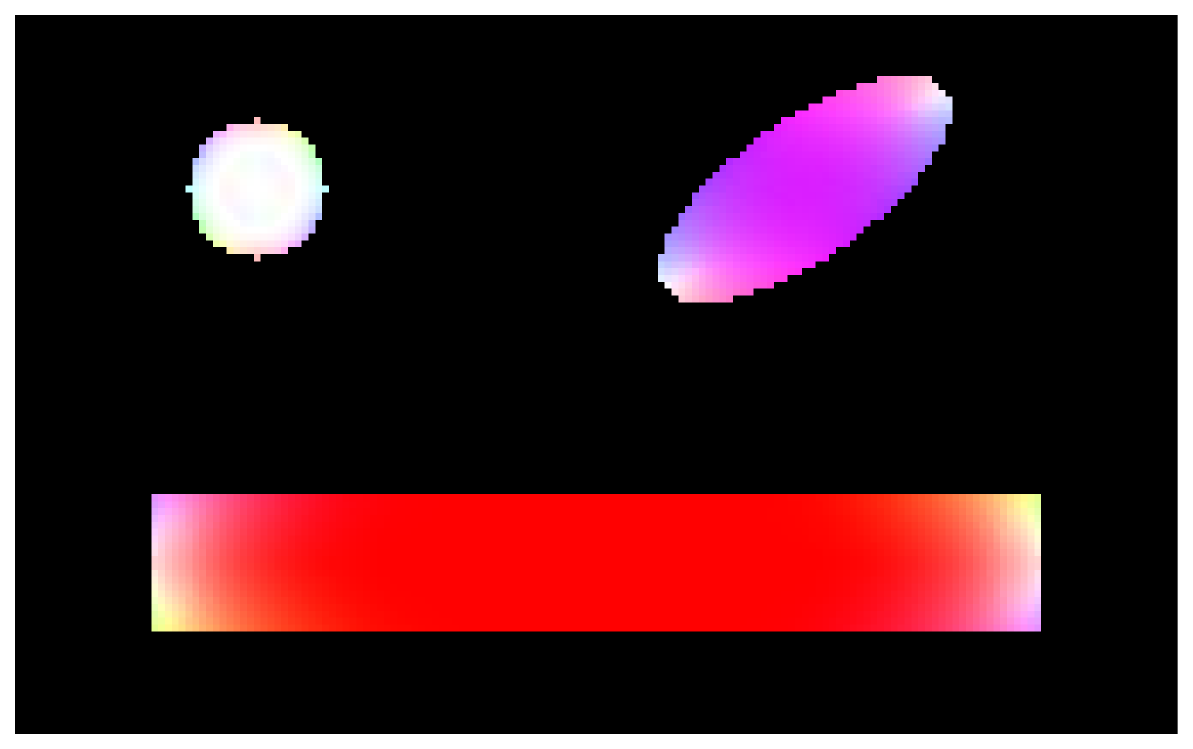}
    \end{tblr}
    \caption{Anisotropy and orientation of the test image in Fig.~\ref{fig:ex1_sample_image} obtained from an approximately optimal single scale and two scale-space approaches. Orientation is represented as color, where each angle in $(0,\pi]$ has a unique hue value (horizontal\,--\,red, $45^\circ$\,--\,purple). We use whiteness as anisotropy in the orientation plot as it affects the relevance of the orientation values. Given a relatively uniform feature size, the single scale result serves as an approximate baseline for the other two methods. Results exclude areas outside the features. Both scale-space methods return generally correct results, but our scale-space is more accurate for the circular feature and along the edges of features.}
    \label{fig:2d_art}
\end{figure}

Based on anisotropy and orientation, the most elementary measures extracted from structure tensor, we compare our scale-space method to the conventional version~\cite{Lindeberg1994}, see Fig.~\ref{fig:2d_art}. The difference between the methods is not very big for this simple problem. The single scale result is very accurate but affected by the strong smoothing that is necessary without the ring filter. In the baseline scale-space, the inaccuracy is primarily visible on the edges and corners of the feature. The proposed method returns a much cleaner anisotropy map compared especially to the baseline scale-space method, but also to the single scale result - data in the circle is more uniform, and the tips of the anisotropic features are correctly found to be locally isotropic. Although less visible, similar conclusions can be drawn based on the orientation data.  

\begin{figure}[!t]
     \begin{subfigure}[b]{0.49\columnwidth}
         \centering
         \includegraphics[width=\textwidth]{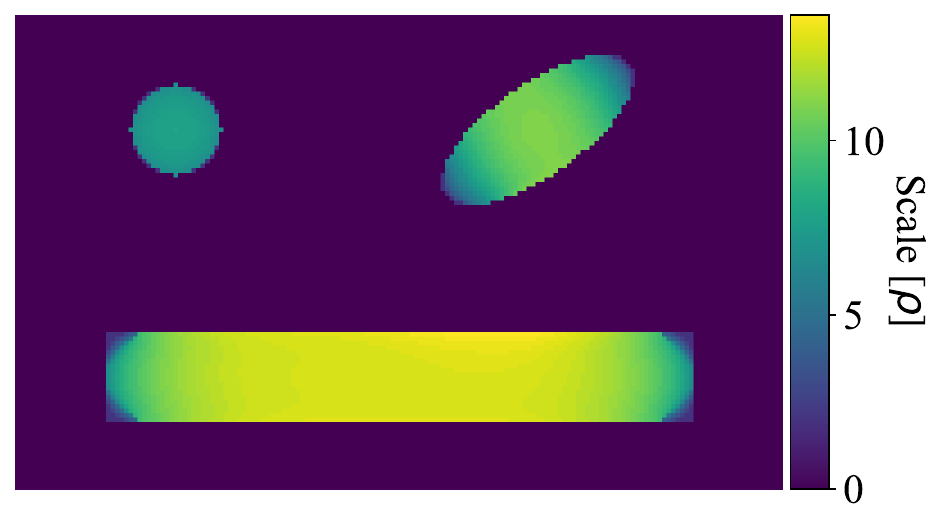}
         \caption{Baseline scale map.}\quad
         \label{fig:8a}
     \end{subfigure}
          \begin{subfigure}[b]{0.49\columnwidth}
         \centering
         \includegraphics[width=\textwidth]{figures/method/scaleEx_norm_im.pdf}
         \caption{Proposed scale map after correction.}
         \label{fig:8b}
     \end{subfigure}
\caption{Comparison of scale maps between the baseline and our method. The discriminative power is similar but less accurate and strongly smoothed out with the baseline method. Due to the lack of scale-feature size correlation, the baseline provides much less insight into the dimensions of the features.}
\label{fig:2Dart_scale_oldScale}
\end{figure}

The biggest difference between the two scale-space methods appears in the scale information they return (Fig.~\ref{fig:2Dart_scale_oldScale}). The existing method has clear discriminative power, enabled mainly by the strong smoothing of the second Gaussian filter, but its scale map is much less useful, primarily because its scale values cannot be correlated to an exact feature size. The main reason for this is the different scale normalization approach, but also the response distortion caused by smoothing. These two problems make it impossible to apply the same scale correction scheme as in the proposed method and cause distortions of the values at the corners of the features.

\begin{figure}[!t]
     \centering
     \begin{subfigure}[b]{0.22\textwidth}
         \centering
         \includegraphics[width=\textwidth]{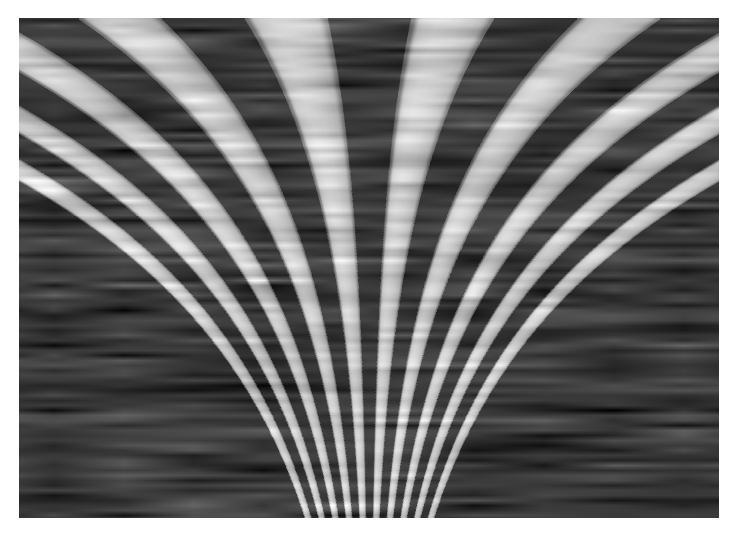}
         \caption{Input image.}
         \label{fig:fiberWarpedAnis}
     \end{subfigure}
     \begin{subfigure}[b]{0.24\textwidth}
         \centering
         \includegraphics[width=\textwidth]{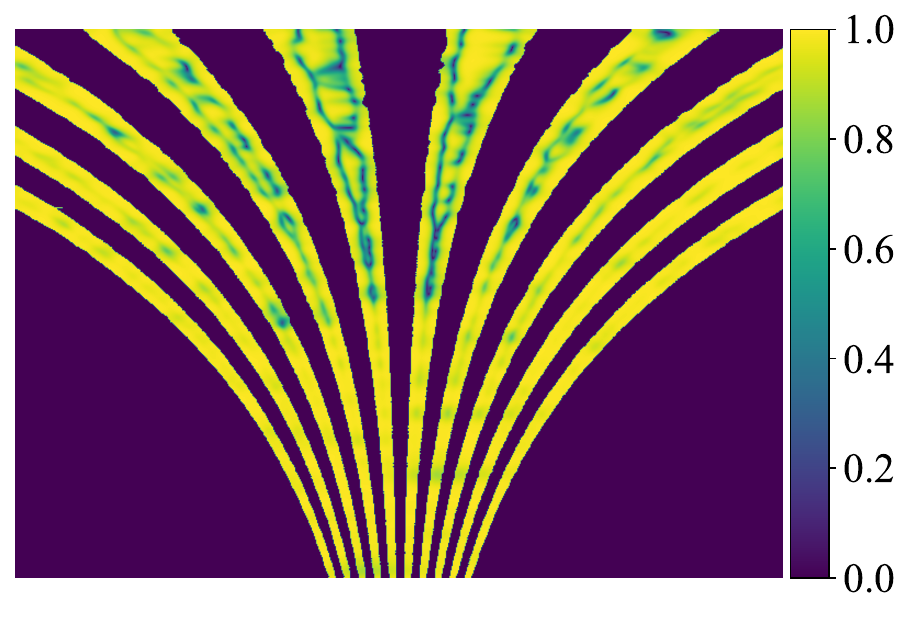}
         \caption{Anisotropy at scale $\sigma=2$.}
         \label{fig:warpAnis_scale2_anis}
     \end{subfigure}
     \begin{subfigure}[b]{0.23\textwidth}
         \includegraphics[width=\textwidth]{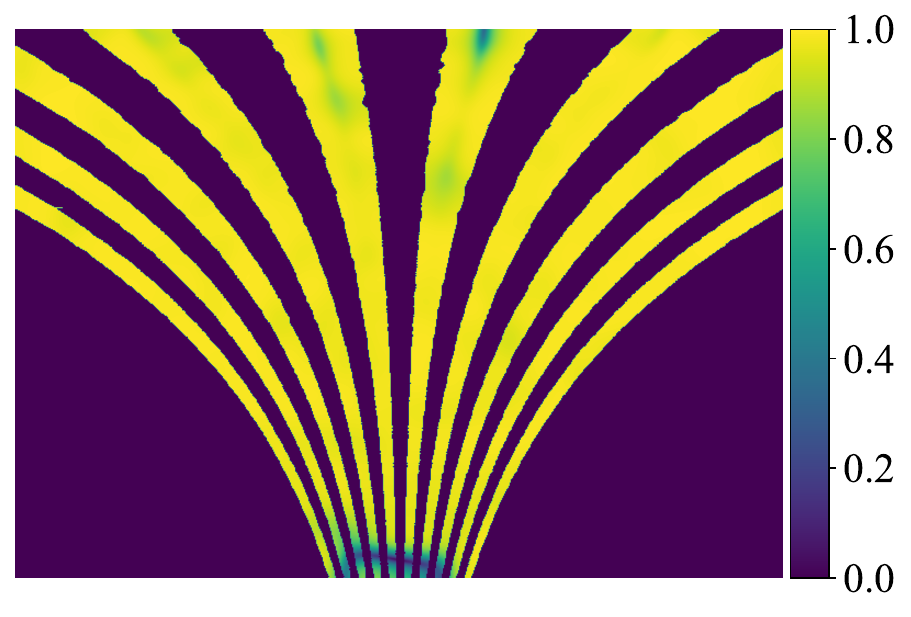}
         \caption{Anisotropy at scale $\sigma=6$.}
         \label{fig:warpAnis_scale6_anis}
     \end{subfigure}
     \begin{subfigure}[b]{0.23\textwidth}
         \centering
         \includegraphics[width=\textwidth]{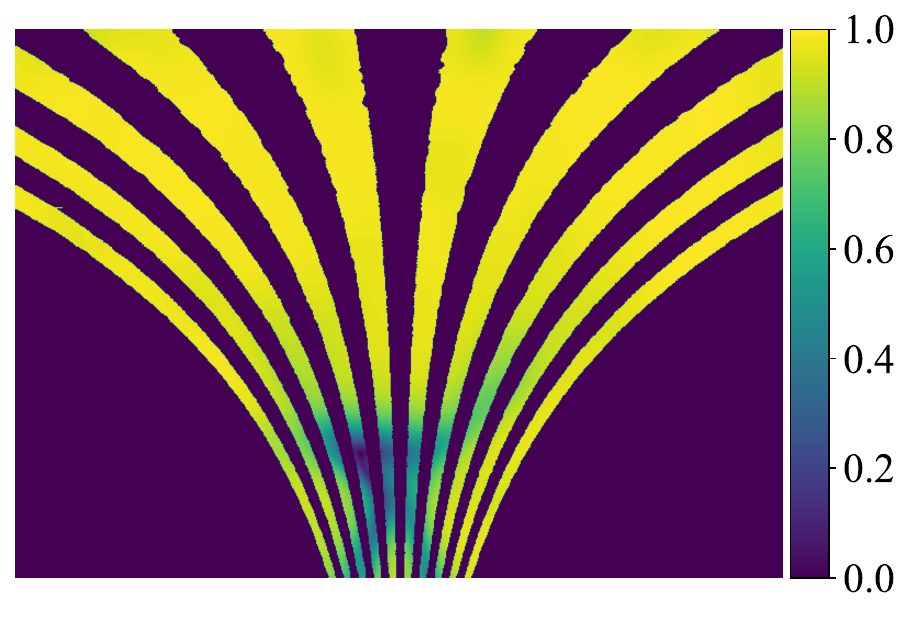}
         \caption{Anisotropy at scale $\sigma=12$.}
         \label{fig:warpAnis_scale12_anis}
     \end{subfigure}
    \caption{Test image consisting of vertical lines with increasing size and with anisotropic noise, and its anisotropy calculated in three scales. In all three cases, $\rho = 2\sigma$. Due to a variable feature size and anisotropic noise, getting an accurate anisotropy measurement with a single scale is not possible.}
    \label{fig:fiberWarped}
\end{figure}

To demonstrate an application of our method, we use another artificial image that consists of vertical lines gradually increasing in size, see Fig.~\ref{fig:fiberWarpedAnis}. Assuming that the single scale approach distorts the result when the filters do not align well, using only a single scale should result in a visible local decrease of anisotropy for these very anisotropic features. To better demonstrate this effect, an anisotropic noise is added to the image. Without this (or with isotropic noise), the smoothing coming from the integrating Gaussian would help cover a wide range of feature sizes, making it necessary to grow the range of the line widths. Even though this exact noise is not common in real cases, it simulates a very real situation where strong features are placed around other, weaker features that should be filtered out.

\begin{figure}[!t]
     \centering
     \begin{subfigure}[b]{0.23\textwidth}
         \centering
         \includegraphics[width=\textwidth]{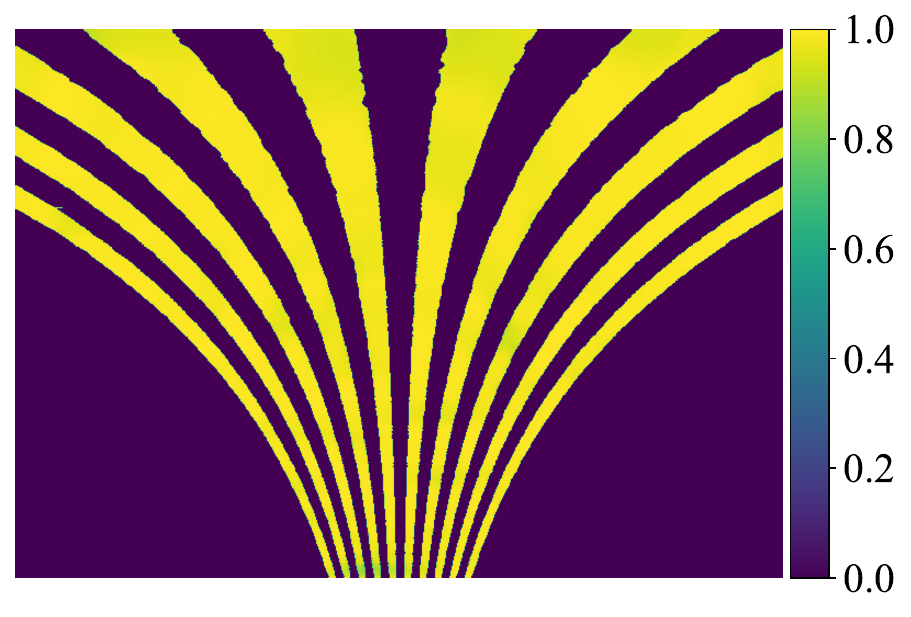}
         \caption{Anisotropy.}
         \label{fig:warpAnis_scaleSpace_anis}
     \end{subfigure}
     \begin{subfigure}[b]{0.24\textwidth}
         \centering
         \includegraphics[width=\textwidth]{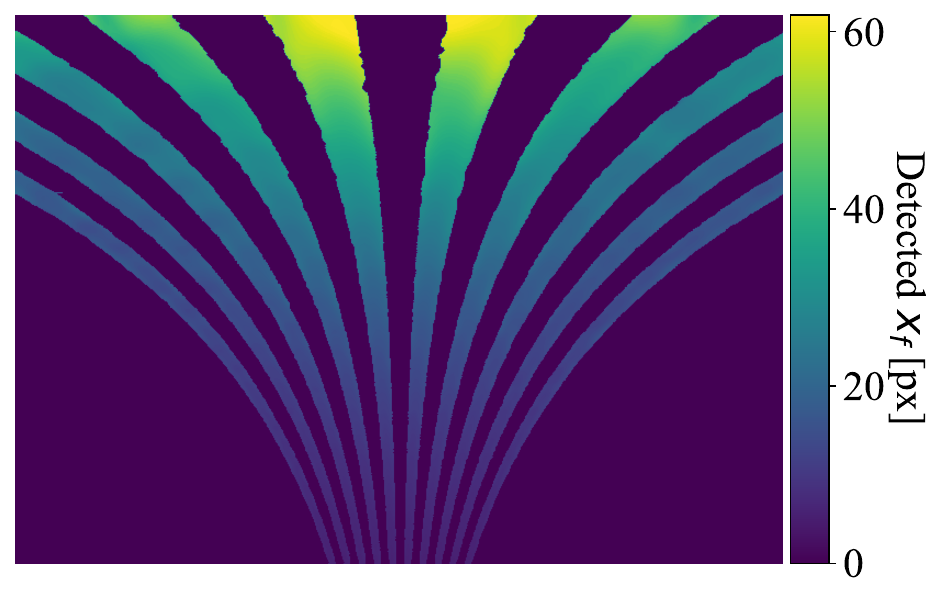}
         \caption{Feature width.}
         \label{fig:warpAnis_scaleSpace_scale}
     \end{subfigure}
    \caption{Anisotropy and scale using proposed scale-space approach calculated on a test image from Fig. \ref{fig:fiberWarpedAnis}. The anisotropy measure is accurate for the whole image and there is a smooth feature size progression.}
    \label{fig:fiberWarpedScale}
\end{figure}

The extracted anisotropy measure confirms that a single scale result does not correctly represent the structure. This distortion is especially visible for badly chosen scales (Figs.~\ref{fig:warpAnis_scale2_anis} and \ref{fig:warpAnis_scale12_anis}), but even after experimentally choosing the best possible scale (Fig.~\ref{fig:warpAnis_scale6_anis}), the distortion is still noticeable. The scale-space result returns strong anisotropy for all segments and provides a smoothly transitioning scale map (Fig.~\ref{fig:fiberWarpedScale}).

\subsection{Real 3D data} \label{sec:exp_real3D}

The most realistic test of the method comes from investigating a scanned 3D dataset. Even though it is much simpler to describe the method in 2D, our current interest is primarily the use of the structure tensor in 3D cases. This is because a 2D image structure is rarely an accurate representation of the analyzed object --- both due to the missing dimension and the pixel intensities being dependent on the imaging conditions, such as lighting or shade. For these experiments, we use $\mu$CT scans of two objects: glass fiber bundles and mozzarella cheese microstructure.

The first 3D image contains a small subsection from a high-resolution scan of a composite material used in wind turbine blade manufacturing \cite{jespersen2016a}. The selected region contains two distinct fiber groups: one composed of thin fibers aligned along the Z-axis and another of wider fibers aligned along the Y-axis. This straightforward but realistic 3D example enables a visual assessment of the method’s accuracy in analyzing fiber orientation.

\begin{figure*}[!t]
     \centering
     \begin{subfigure}[b]{0.24\textwidth}
         \centering
         \includegraphics[width=\textwidth]{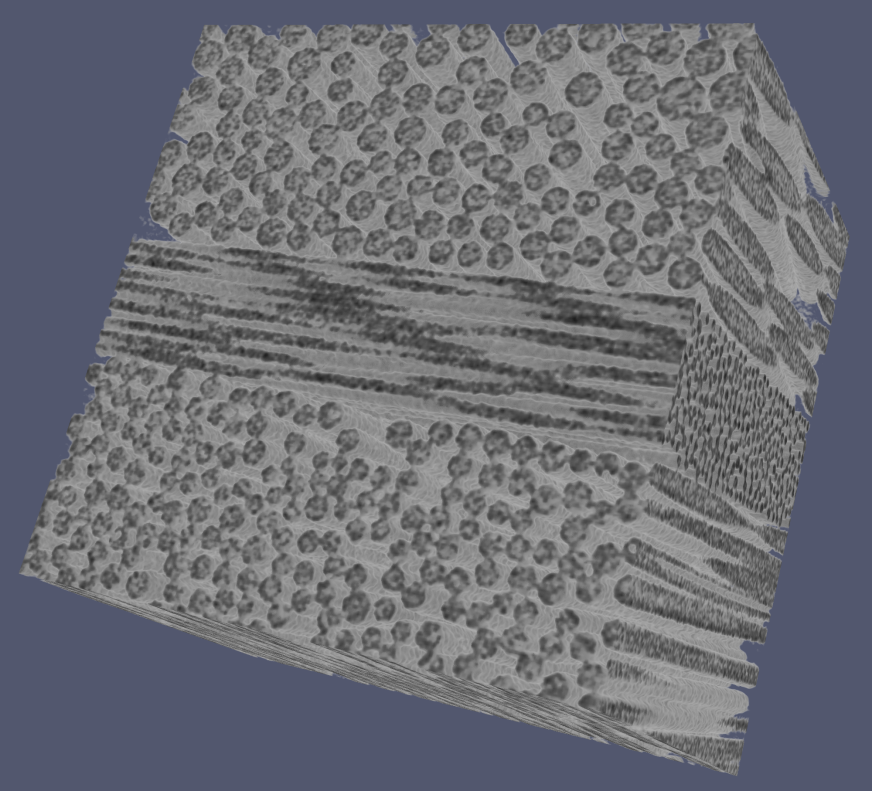}
         \caption{Input volume.}
         \label{fig:fiber3d_raw}
     \end{subfigure}
     \begin{subfigure}[b]{0.24\textwidth}
         \centering
         \includegraphics[width=\textwidth]{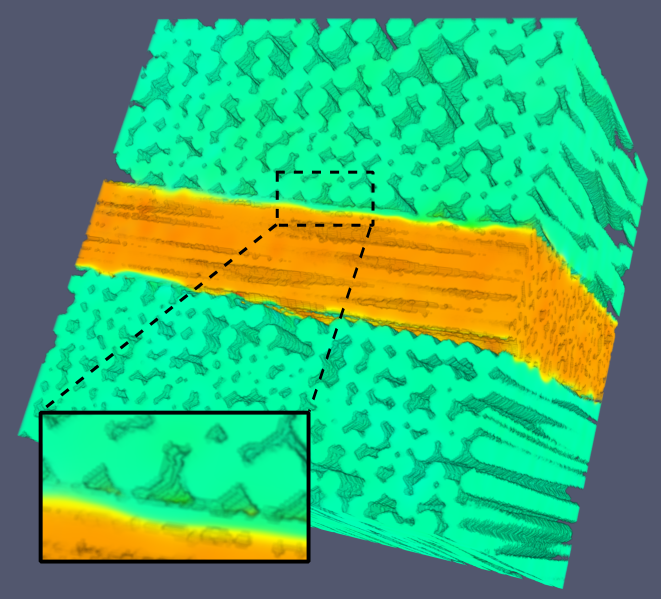}
         \caption{Single scale, baseline.}
         \label{fig:fiber3d_single_no_ring}
     \end{subfigure}
     \begin{subfigure}[b]{0.24\textwidth}
         \centering
         \includegraphics[width=\textwidth]{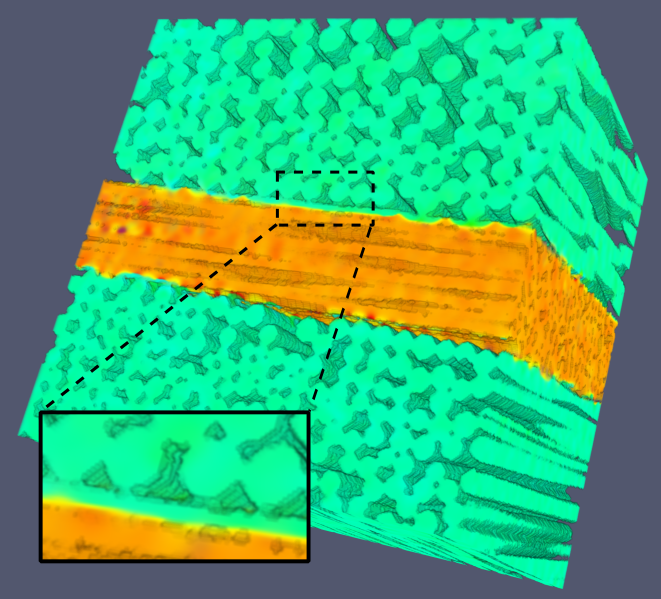}
         \caption{Single scale, with ring filter.}
         \label{fig:fiber3d_single_ring}
     \end{subfigure}
     \begin{subfigure}[b]{0.24\textwidth}
         \centering
         \includegraphics[width=\textwidth]{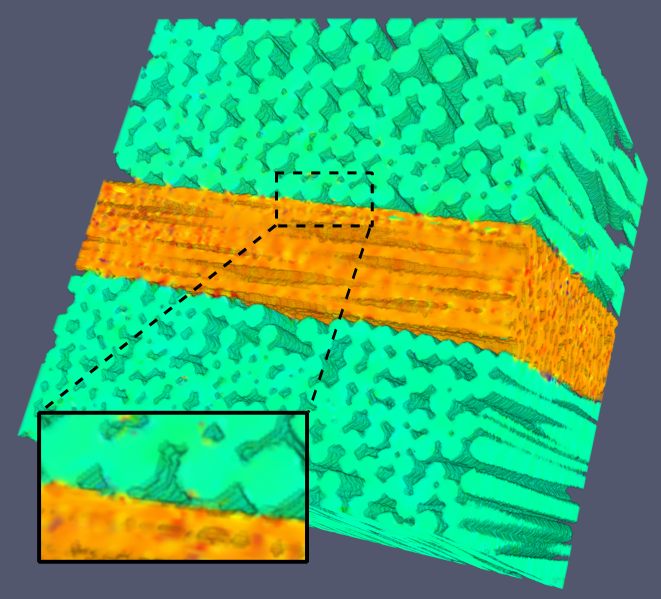}
         \caption{Proposed scale-space.}
         \label{fig:fiber3d_scale_space}
     \end{subfigure}
    \caption{Fiber orientation analysis result using three instances of the structure tensor method. All methods return generally correct results, but the proposed scale-space provides the highest accuracy in more complex areas. All volumes are visualized with the background removed. The colors map the orientation space as demonstrated in Fig. \ref{fig:orientSphere}. }
    \label{fig:fiber3d}
\end{figure*}

We evaluate three variations of the structure tensor method: a baseline single-scale calculation (using parameters that are optimal for the wider fibers), a single scale with the ring filter, and the proposed scale-space approach. Orientation results clearly highlight the accuracy of the proposed method (Fig. \ref{fig:fiber3d}). All three instances returned generally correct orientation values, but both single scale attempts demonstrated high inaccuracies on the edges between the two fiber types. Furthermore, the baseline method returned a highly-smoothed orientation description, effectively removing the details that can be found in both instances that used the proposed modifications. 

In the second experiment, we are using a CT scan of mozzarella cheese, which is known to have a distinct anisotropic structure that affects its properties~\cite{Bast2015, Feng2021, Feng2023}. Because of this structure, it is interesting to quantify and analyze the anisotropy and orientation of the cheese fibers in relation to the functional properties of the cheese. 

\begin{figure*}[!t]
     \centering
     \begin{subfigure}[b]{0.22\textwidth}
         \centering
         \includegraphics[width=\textwidth]{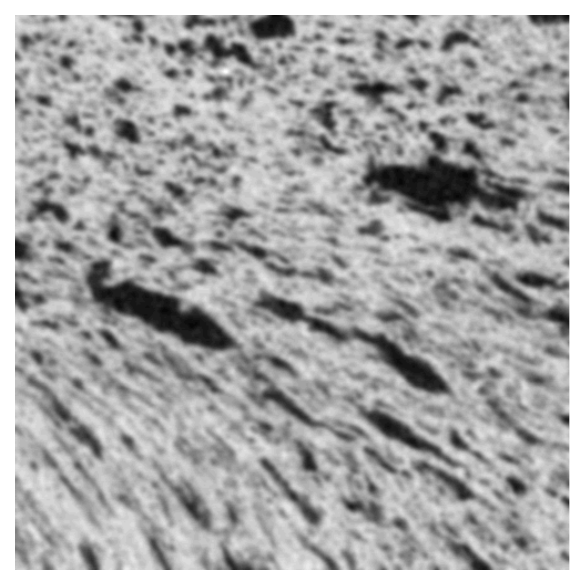}
         \caption{Slice from a CT scan.}
         \label{fig:cheeseSlice}
     \end{subfigure}
     \begin{subfigure}[b]{0.22\textwidth}
         \centering
         \includegraphics[width=\textwidth]{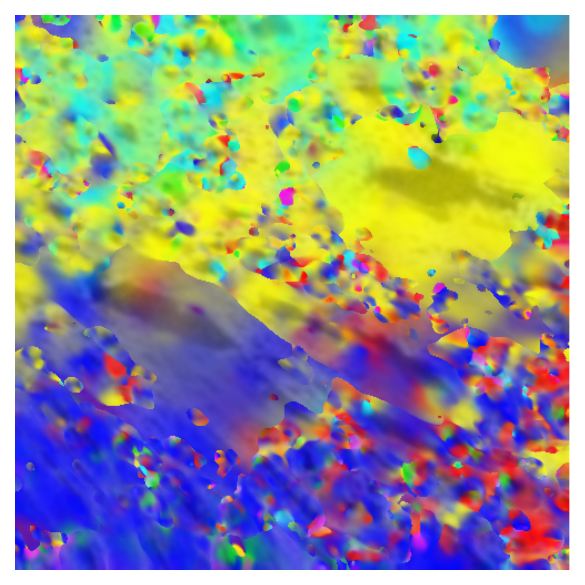}
         \caption{Orientation.}
         \label{fig:cheeseOrientSlice}
     \end{subfigure}
     \begin{subfigure}[b]{0.27\textwidth}
         \centering
         \includegraphics[width=\textwidth]{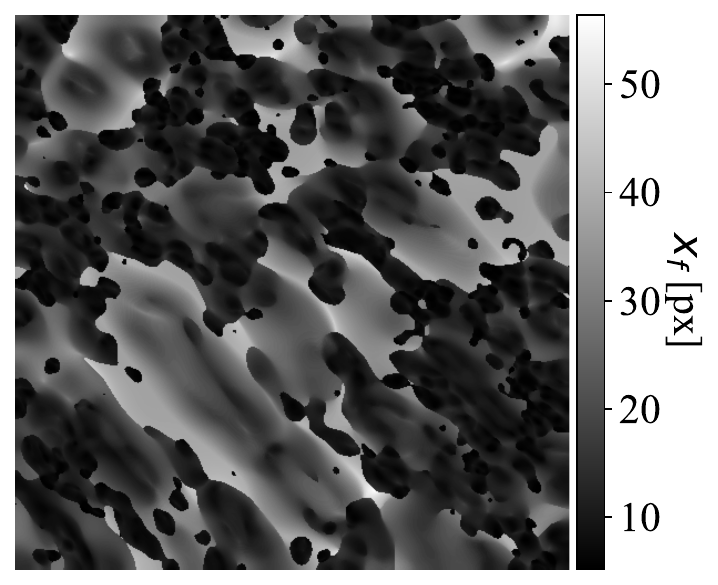}
         \caption{Feature width.}
         \label{fig:cheeseScaleSlice}
     \end{subfigure}
     \begin{subfigure}[b]{0.24\textwidth}
         \centering
         \includegraphics[width=\textwidth]{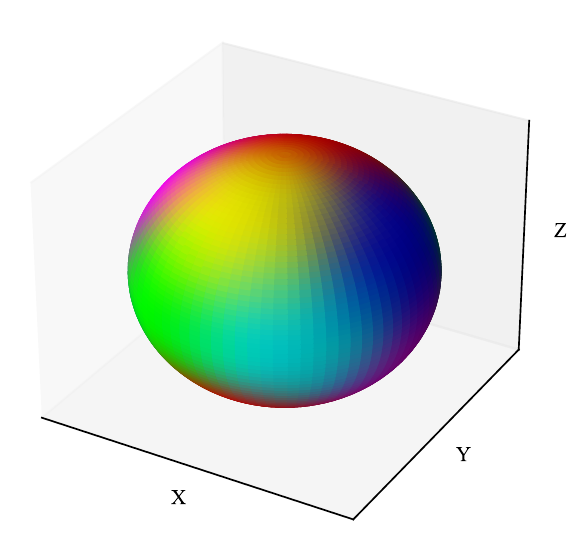}
         \caption{Visualization of the color-to-orientation mapping.}
         \label{fig:orientSphere}
    \end{subfigure}
    \caption{Slice from a CT scan of mozzarella cheese and its feature width map and orientation after performing scale-space structure tensor analysis. At least two distinct orientations are visible, with tightly-placed features found across a wide range of sizes.}
    \label{fig:cheese}
\end{figure*}

A CT scan of mozzarella serves as a good example of the necessity of a scale-space solution. The features in the volume are tightly packed, have strongly varying sizes, and often change directions (see Fig.~\ref{fig:cheeseSlice}). In such cases, it can be very hard to choose a correct scale value, which is a problem that was not demonstrated in the previous experiments. Oftentimes, the general result of many different scales seems correct and it can be hard to find a visible difference between two scales without a pixel-wise comparison of each value. Despite them appearing similar, they apply a strong smoothing that distorts local results, especially with tightly packed features of varying orientation and size. As a result, every single scale calculation seems both equally correct and incorrect, and there is no accurate way of judging its quality.

To artificially demonstrate the level of inaccuracy caused by the single scale, we perform a structure tensor calculation on an original volume and a volume rescaled to half its original size. This approach simulates an application of the method on features that have a varying size; with the rescaling serving as a way to control this size variation. It is then possible to compare a pixel-wise difference in orientation angle for both these volumes and quantify the accuracies of the scale-space and single scale approaches. 

We compare a set of single-scale calculations using the baseline structure tensor, as well as using our version with a ring filter instead of the Gaussian integration. 

The orientation difference results in Table~\ref{tab:err_stat} confirm that even though in some cases a single scale calculation can cover a wide range of feature sizes, it can also cause a considerable inaccuracy in cases that are more complex. The mean and standard deviation of all single scale results are significantly higher than our scale-space approach which demonstrates strong scale invariance. This difference in accuracy is especially noticeable in comparison to the ring filter version, which suggests that the strong smoothing decreases the local angle differences, distorting the actual accuracy of the baseline.

\begin{figure}[!t]
     \centering
     \begin{subfigure}[b]{0.23\textwidth}
         \centering
         \includegraphics[width=\textwidth]{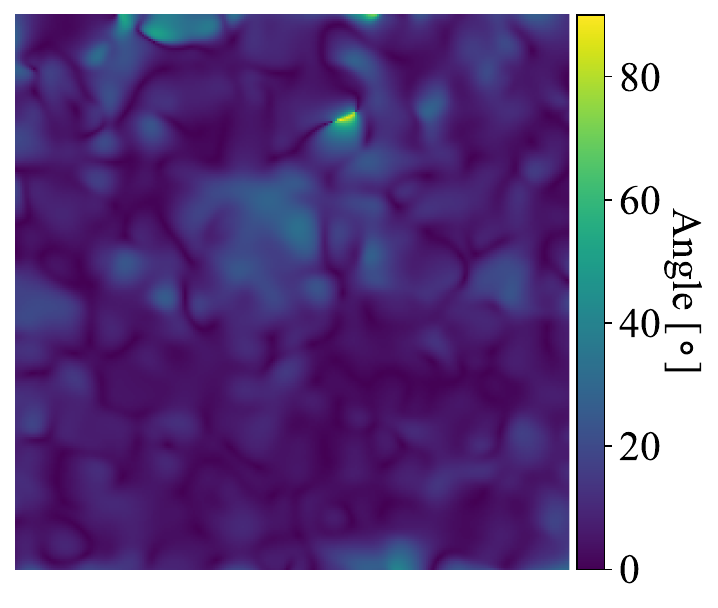}
         \caption{$\sigma=6, \rho=2\sigma$.}
         \label{fig:cheeseOrientDiff1Scalepng}
     \end{subfigure}
     \begin{subfigure}[b]{0.23\textwidth}
         \centering
         \includegraphics[width=\textwidth]{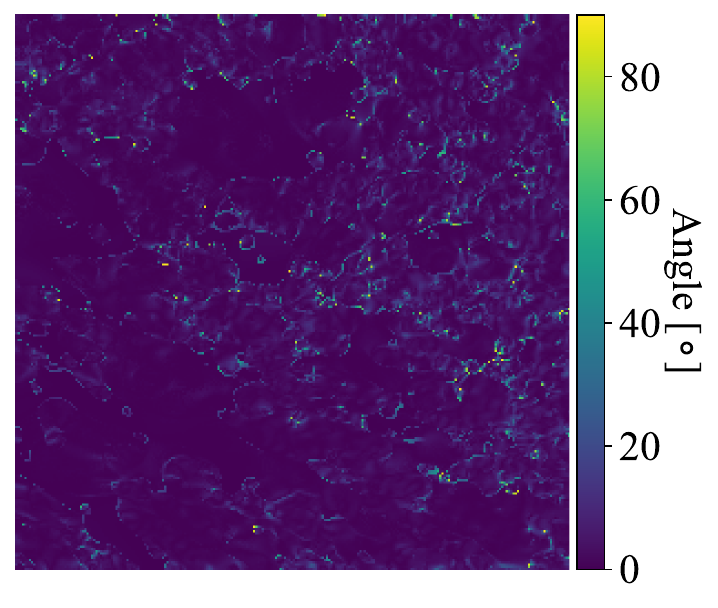}
         \caption{Scale-space.}
         \label{fig:cheeseOrientDiffScaleSpace}
     \end{subfigure}
    \caption{Visualisation of the character of orientation difference errors for the single scale and scale-space approach, based on an example volume slice. In the single scale, the errors are higher and do not have a very specific structure, whereas in the scale-space they are much lower and are placed on the edges of features. Error values are in degrees.}
    \label{fig:cheese_orient_diff}
\end{figure}

Investigation of the orientation difference on an example slice (Fig.~\ref{fig:cheese_orient_diff}) shows a relatively unstructured spread of orientation inaccuracy for the single scale data, whereas for the scale-space approach, it is primarily located on edges of features, which suggests that they primarily come from the information loss caused by rescaling.

\begin{table}
\centering
\caption{Mean and standard deviation of orientation difference between baseline single scale and our scale-space approach to structure tensor calculation. The last column uses the ring filter in place of the Gaussian integrating filter, for the value of $\sigma=6$ as this returned the lowest difference with the baseline method.}
\label{tab:err_stat}
\resizebox{\linewidth}{!}{
\begin{tblr}{
  row{even} = {c},
  row{3} = {c},
  cell{1}{1} = {r=2}{},
  cell{1}{2} = {c=4}{c},
  cell{1}{6} = {c},
  cell{1}{7} = {c},
  vline{2} = {1-4}{0.08em},
  vline{6-7} = {1-4}{0.08em},
  hline{3} = {-}{0.16em},
}
              & $\rho=2\sigma$ &                        &              &               &  scale-space & ring filter \\
              & $\sigma=1$  & $\sigma=3$   & $\uline{\sigma=6}$ & $\sigma=12$ &           (ours)      & $\sigma=6$         \\
\textbf{Mean [$^{\circ}$]} & 20.13         & 11.69 & \uline{10.67}        & 14.52         & 2.68                & 19.51                \\
\textbf{Std [$^{\circ}$]}  & 18.72         & 10.80           & \uline{9.58}         & 12.90         & 5.87                  & 16.65                         
\end{tblr}
}
\end{table}

\subsection{Scale Range Definition Aid}

So far the scale information was primarily used as an aid to demonstrate the accuracy and consistency of the proposed method. In real scenarios, its important use case is to serve as an additional statistic describing an analyzed structure. In addition, the scale-space approach can effectively be used during the execution of the algorithm to adjust the scale range or to post-process the result. 

\begin{figure*}[!t]
     \centering
     \begin{subfigure}[b]{0.48\textwidth}
         \centering
         \includegraphics[width=\textwidth]{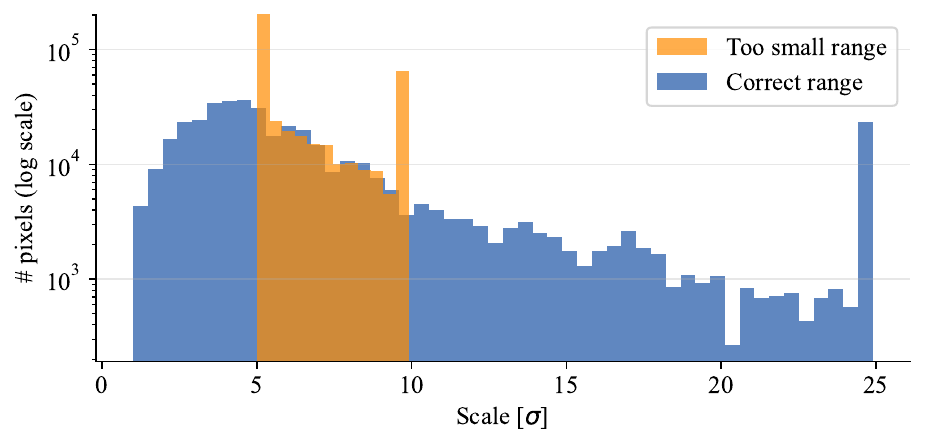}
         \caption{Scale histogram using different scale ranges (before scale correction).}
         \label{fig:2Dmask_hist_badGood_no_corr}
     \end{subfigure}
     \begin{subfigure}[b]{0.48\textwidth}
         \includegraphics[width=\textwidth]{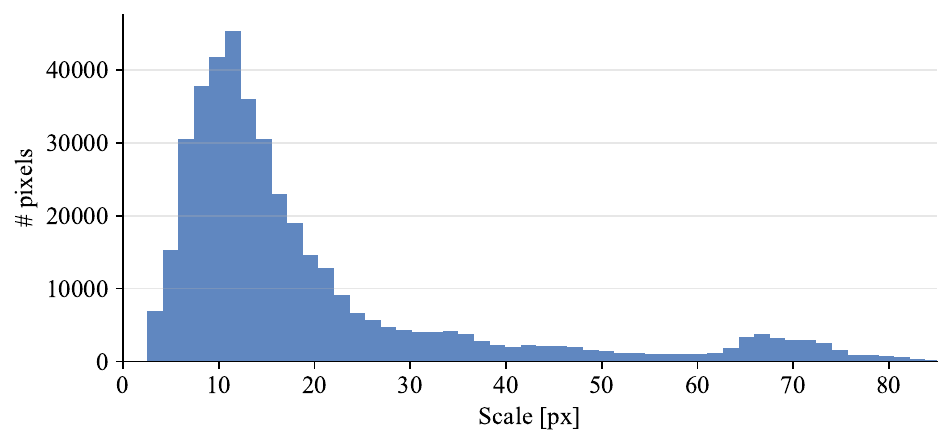}
         \caption{Scale histogram using correct scale range (after scale correction). \quad}
         \label{fig:2Dmask_hist_good_with_corr}
     \end{subfigure}
    \caption{Scale histograms for an image containing a structure with uniform feature width. The histogram can aid in choosing the correct scale range (a), or be used in post-processing to remove distorted results that were assigned a very high scale (b). The scale correction described in Section \ref{sec:mod_scale_corr} has a smoothing effect on the highest scale peak, so it is not applied in (a).}
    \label{fig:2Dmask_hist}
\end{figure*}

If the scale histogram returns a strong peak at the currently lowest or highest scale, we expand the range in the corresponding direction to fully cover the range of present feature sizes and increase the accuracy of the result (Fig.~\ref{fig:2Dmask_hist_badGood_no_corr}). If a strong peak is found at a very low scale (1-3 pixels), it can mean that the image is too noisy, or its resolution is too low. If it is not possible to acquire a better quality image, the simplest solution is to smoothen it before running the structure tensor analysis (which also had to be done in the mozzarella example from Fig.~\ref{fig:cheese}). It is also worth noting that in most cases the high-scale peak will only reach 0 when the scale reaches the image width, and the convolutions with big filters are computationally expensive, so it is not always practical to continue extending the upper scale range. 

Assuming that some prior information is available about the size of analyzed features, it is also possible to use the scale information to filter out areas with scales that are considerably bigger. The example in Fig.~\ref{fig:2Dmask_hist_good_with_corr} shows a scale histogram for such a 2D image with uniform feature size. A clear Gaussian-like peak appears at a feature width of 12\,px corresponding to the expected feature size. The rest of the histogram spans a wide range of bigger values that occur very rarely. They are most likely found in areas where features are not visible enough or lie tightly next to or on top of each other. It is expected that the structural information in these areas does not reflect the actual structure of the imaged objects, and with the scale-space approach, it is easily discarded.

\section {Discussion}

The primary limitation of most scale-space-based methods is execution time. Assuming the filter size of $k$  pixels, the amount of necessary operations is proportional to $k^2$, and this is only for a single dimension. For our method, this limitation is magnified, as the ring filter requires one more convolution operation in each dimension as compared to the baseline integrating Gaussian. 

A set of known properties, such as the separability of the Gaussian kernel~\cite[p.~114]{Burger2009} and its semi-group structure~\cite{Lindeberg1994} can help reduce the computational complexity. The first one of these can be used in the proposed method and was mentioned in Sec.~\ref{sec:ring_filter}. The second property in practice allows for fast calculation of each consecutive scale by reusing the result from the previous one. This method fails in combination with the structure tensor, due to the pixel-wise multiplication of the gradients, which breaks the chain of Gaussian convolutions. 

As a result of these limitations, the structure tensor is not the most practical algorithm for the application of scale-space, especially in combination with large images and wide features. Serving as an example, the execution time for the 500-pixel wide cube from the cheese CT scan introduced in Sec.~\ref{sec:exp_real3D} was approximately 80 minutes for $\sigma$ ranging from 1 to 14 with a step of 1. For a step size of 0.1 (used for smooth visualization), the runtime was 19 hours. Such a result does not deem the method unusable but definitely slows down the data analysis process. It is however important to note that the discussed implementation is single-threaded only, parallelization of structure tensor has been demonstrated and provides a significant increase in efficiency \cite{Jeppesen2021}.

The slow execution time is heavily compensated by the simplicity of the method's application. Any scale-space approach removes the need for manual parameter choice, but in this case, it is especially noticeable. One of the most vaguely defined properties in the conventional method was the ratio between the $\sigma$ and $\rho$ parameters. The proposed method replaces the integrating step with the ring filter, removing the need for the $\rho$ parameter. As a result, from the user's perspective, the complex two-parameter optimization task is simplified to defining a range of values for one parameter, which can be even further simplified, or possibly automated based on the scale histogram.

On the low level, the biggest uncertainty is brought by the design of the newly introduced ring filter. It causes the filter slopes to be uneven, which can result in small inaccuracies (although this effect has not been observed in practice). Additionally, the relation of ring width to its radius has not been researched and is likely not optimal for the given task. We decided not to investigate this further based on the assumption that the simplicity of the filter (and subsequent computational speed) was more important than optimizing its exact shape. 

Another limitation is found in the observation that extracted scale information is only accurate at the skeletons of the features, which was also mentioned in Sec.~\ref{sec:mod_scale_corr}. It is however important to note that there is no standard way of defining the correct local width at each pixel within a feature. It is also unrealistic to expect a single scale value within the whole feature, directly from a convolution-based method, especially with real and noisy data. Nevertheless, the extracted information cannot truly serve as an accurate, dense scale map. It has to be interpreted as an apparent feature size information that on average returns relatively accurate results. 

The final limitation often discussed in conjunction with scale-space solutions is the sensitivity to noise. For our method, the main source of this sensitivity comes from the alignment of the derivative filter to a feature. Our unique approach means that the method may not be perfectly accurate in cases where the feature's edge is smooth, but this inaccuracy will always be a small fraction in relation to the whole feature width. Despite those theoretical limitations, we have experimentally demonstrated that our method has a high resistance to noise (Figs.~\ref{fig:fiberWarped} and \ref{fig:fiberWarpedScale}).

\section {Conclusion}

We have introduced a novel approach for computing a first order structure tensor scale-space. We have provided a mathematical basis for the proposed modifications and benchmarked them against existing widely recognised single scale and scale-space solutions, both for artificial 2D data and real 3D data. 

Our results demonstrate significant improvements as compared with the tested baselines. In simple cases with features of uniform size, the accuracy of our method is similar to that of a correctly defined single scale calculation, with the added benefit of a weaker smoothing effect due to the introduced ring filter. In comparison with the baseline scale-space method, our algorithm demonstrates higher accuracy of the extracted anisotropy, as well as increased reliability and usability of the resulting scale information, bridging the gap between filter and feature size.

In complex cases with variable feature size, our method demonstrates scale invariance and resistance to noise, providing highly accurate anisotropy and orientation data, while also enabling a simple scale range choice.

Limitations of the method center primarily around the high execution time of scale-space solutions, magnified in this case by the inherent properties of the structure tensor method, as well as the additional complexity introduced with the ring filter. There is also a possibility of small inaccuracies introduced by the design decisions in the scale normalization and ring filter.

In summary, our method is an accurate, scale-invariant tool for structural analysis, both in 2D and 3D, requiring minimal user input while ensuring the use of optimal parameters for a given feature. The method further provides the user with a scale map that can serve as a rudimentary measure of object width and be utilized in downstream statistical analysis. An implementation of our methods is available at \url{https://github.com/PaPieta/stss}. The presented results can be recreated on an associated Code Ocean capsule: \url{https://codeocean.com/capsule/8105965/tree}. All software development was performed in Python, using NumPy and SciPy libraries.

\vfill

\begin{appendices}
\section {\break Derivation of Eq.\lowercase{~\ref{eq:gamma_ratio}}} \label{App:proof_eq_norm_1}

Starting from Eq.~\ref{eq:rect_conv} and calculating the integrals within the given range, we have:
\begin{align}
    &P(x_f, \sigma, \gamma) \notag\\
    & = 2\sigma^\gamma \Big(g(0, \sigma) - [ g(x_f, \sigma) - g(0, \sigma) ] - g(-x_f, \sigma) \Big) \notag \\
    & = 2\sigma^\gamma \Big(2g(0, \sigma) - 2g(x_f, \sigma)\Big)\\
    & = \frac{4\sigma^{\gamma}}{\sigma\sqrt{2\pi}} \Big(1 - e^{\frac{-x_f^2}{ 2\sigma^2}}\Big) \propto \sigma^{\gamma-1} \Big(1 -  e^{\frac{-x_f^2}{ 2\sigma^2}}\Big) .
\end{align}
Taking the gradients of the filter response $P$ with respect to the standard deviation $\sigma$,
\begin{align}
    \frac{\partial P}{\partial \sigma} &\propto (\gamma-1)\sigma^{\gamma-2} \Big(1 - e^{\frac{-x_f^2}{ 2\sigma^2}} \Big) - \sigma^{\gamma-1} \Big(x_f^2 e^{\frac{-x_f^2}{ 2\sigma^2}} \Big) \sigma^{-3} \notag \\
    & = (\gamma-1)\sigma^{\gamma-2} \Big(1 - e^{\frac{-x_f^2}{ 2\sigma^2}} \Big) - \sigma^{\gamma-2} \Big(\frac{x_f^2}{\sigma^{2}} e^{\frac{-x_f^2}{ 2\sigma^2}} \Big) .
\end{align}
Setting $t = \frac{\sigma}{x_f}$,
\begin{align}
    \frac{\partial P}{\partial \sigma} & \propto (\gamma-1)\sigma^{\gamma-2} \Big(1 - e^{\frac{-1}{2t^2}} \Big) - \sigma^{\gamma-2} \Big(\frac{1}{t^2} e^{\frac{-1}{2t^2}} \Big) .
\end{align}
Setting the gradient $\frac{\partial P}{\partial \sigma}=0$ for $\sigma > 0$,
\begin{align}
    &(\gamma-1) \Big(1 - e^{\frac{-1}{2t^2}} \Big) -  \Big(\frac{1}{t^2} e^{\frac{-1}{2t^2}}\Big) = 0 \\
    \gamma  &= \frac{e^{\frac{-1}{2t^2}}}{t^2 \Big(1 - e^{\frac{-1}{2t^2}} \Big)} +1 \label{App_eq:gamma_v2} =\frac{1}{t^2 \Big(e^{\frac{1}{2t^2}} -1 \Big)} +1 \,.
\end{align}

\section{\break Derivation of Eq.\lowercase{~\ref{eq:t_gamma_ratio}}} \label{App:proof_eq_norm_2}

Starting by rearranging Eq.~\ref{App_eq:gamma_v2}:
\begin{align}
    &\frac{1}{\gamma-1} = t^2\Big(e^{\frac{1}{2t^2}}-1\Big) \\
    &\frac{1}{t^2} = e^{\frac{1}{2t^2}}(\gamma-1) - \gamma + 1 \\
    &\Big(\frac{1}{t^2} + \gamma - 1\Big) e^{-\frac{1}{2t^2}} = \gamma -1 \,.
\end{align}
Multiplying both sides by $-\frac{1}{2}e^{-\frac{\gamma-1}{2}}$,
\begin{align}
&-\Big(\frac{1}{2t^2} + \frac{\gamma - 1}{2}\Big)e^{-\Big(\frac{1}{2t^2} + \frac{\gamma - 1}{2}\Big)} = -\frac{\gamma-1}{2} e^{-\frac{\gamma-1}{2}} \,.
\end{align}
Applying the definition of Lambert $W$ function $ye^y = x \Leftrightarrow y = W(x)$. Here $x$ is negative (assuming $\gamma > 1$), so $W_{-1}$ branch is used, returning real values for $x \geq -\frac{1}{e} \Leftrightarrow \gamma \leq 3$.
\begin{align}
&W_{-1}\Big(\frac{1-\gamma}{2} e^{\frac{1-\gamma}{2}}\Big) = -\frac{1}{2t^2} + \frac{1-\gamma}{2} \,.
\end{align}
Solving for $t$ (with the assumption that $t>0$):
\begin{align}
&\frac{1}{t^2} = 1-\gamma - 2W_{-1}\Big(\frac{1-\gamma}{2} e^{\frac{1-\gamma}{2}}\Big)\\
& t = \frac{1}{\sqrt{1-\gamma - 2W_{-1}\Big(\frac{1-\gamma}{2} e^{\frac{1-\gamma}{2}}\Big)}} \,.
\end{align}

\section{\break Derivation of Eq.\lowercase{~\ref{eq:ring_filter_max}}} \label{App:proof_eq_ring}
Starting from the derivative of Eq.~\ref{eq:ring_filter}:
\begin{align}
\frac{\partial R}{\partial x} &= \frac{-x}{\sigma^2} e^{\frac{-x^2}{2\sigma^2}} + \frac{x}{\sigma^2 k^2} e^{\frac{-x^2}{2\sigma^2 k^2}} \notag \\
&=\frac{-x}{\sigma^2} \Big( e^{\frac{-x^2}{2\sigma^2}} + \frac{1}{k^2} e^{\frac{-x^2}{2\sigma^2 k^2}}\Big)
\end{align}
Setting $\frac{\partial R}{\partial x}=0$ for $\sigma \neq 0$,
\begin{align}
&\frac{-x}{\sigma^2}=0 \quad \lor \quad e^{\frac{-x^2}{2\sigma^2}} - \frac{1}{k^2} e^{\frac{-x^2}{2\sigma^2 k^2}} = 0
\end{align}
Ignoring $x=0$, corresponding to a local minimum:
\begin{align}
& e^{\frac{-x^2}{2\sigma_R^2}} = \frac{1}{k^2} e^{\frac{-x^2}{2\sigma_R^2 k^2}}\\
& \frac{-x^2}{2\sigma_R^2} = \ln\Big(\frac{1}{k^2}\Big) - \frac{x^2}{2\sigma_R^2 k^2} \\
& \frac{x^2}{2\sigma_R^2} - \frac{x^2}{2\sigma_R^2 k^2} = \ln(k^2)  \\
&  x^2\Big(1-\frac{1}{k^2}\Big) = 2\sigma_R^2 \ln(k^2)  \\
& x = \pm \sqrt{\frac{2\sigma_R^2 \ln(k^2)}{1-\frac{1}{k^2}}} = \pm \sigma_R k \sqrt{\frac{2\ln(k^2)}{k^2-1}} \,.
\end{align}

\section{\break Relation between \lowercase{$x_f$} of an anisotropic feature and optimal $\sigma$} \label{App:proof_of_prop_1}

\begin{figure}[!t]
     \centering
     \begin{subfigure}[b]{0.48\linewidth}
         \centering
         \includegraphics[width=\linewidth]{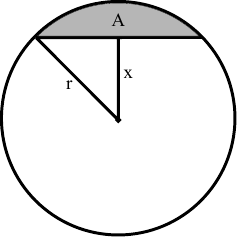}
         \caption{Circular segment.}
         \label{fig:prop1_segment}
     \end{subfigure}
     \begin{subfigure}[b]{0.48\linewidth}
         \includegraphics[width=\linewidth]{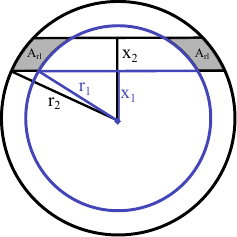}
         \caption{Ring line intersection. \quad}
         \label{fig:prop1_ring_line}
     \end{subfigure}
    \caption{Visualization and variable definition for the circular segment area and ring-line intersection area problems.}
    \label{fig:prop1_ring_line_sector}
\end{figure}

As demonstrated in Fig.~\ref{fig:ex2_derivative_response}, $\mathrm{tr}(\nabla_{0,s})$ of an anisotropic feature after the derivative Gaussian filter consists of two parallel lines. This means that the problem of finding a ring filter radius that returns the strongest response can be simplified to finding the maximum cross-section between one of the two lines and a ring (assuming the ring center lies exactly halfway between the lines). The simplified ring and line functions are defined as (using notation from Fig. \ref{fig:prop1_ring_line}):
\begin{align}
R_{\mathrm{simp}}(x,y)  = & 
    \begin{cases}
      1 & \text{if}\, r \geq r_1 \wedge r < r_2 \\
      0 & \text{otherwise}
    \end{cases}, \\
& \quad r^2= x^2 + y^2 \notag \\
L_{\mathrm{simp}}(x,y)  = & 
    \begin{cases}
      1 & \text{if}\, x \geq x_1 \wedge x < x_2 \\
      0 & \text{otherwise}
    \end{cases}
\end{align}

Using the definition of Full Width at Half Maximum, which is a measure of distribution width, it can be shown that for Gaussian distribution it is proportional to $\sigma$ and equal to:
\begin{align}
G_{\mathrm{FWHM}} = 2\sqrt{2 \ln{2}} \sigma.
\end{align}
With that relation in mind, it can be reasoned that:
\begin{itemize}
    \item The ring filter width is proportional to sigma, as it is constructed by subtracting two Gaussian distributions,
    \item The line width is proportional to sigma, as a convolution of a perfect anisotropic feature (a step edge) and a Gaussian derivative is a Gaussian distribution.
\end{itemize}
In summary, given some scale $\sigma$, ring center radius $r_c$ and distance from the line to ring center $x_c=\frac{1}{2} x_f$, the edge positions of the line and ring are defined as:
\begin{align}
x_1 = x_c - \sigma w_x, \quad x_2 = x_c + \sigma w_x, \notag \\
r_1 = r_c - \sigma w_r, \quad r_2 = r_c + \sigma w_r.
\end{align}
A similar proportionality relation can be drawn for the ring center radius $r_c$ based on Eq.~\ref{eq:ring_filter_max}, resulting in
\begin{align}
r_c = a_r\sigma \quad r_1 = \sigma a_r - \sigma w_r, \quad r_2 = \sigma a_r + \sigma w_r,
\end{align}
where $w_x$, $w_r$ and $a_r$ are some unknown constants.

\begin{prop}
Given above definitions ($R_{\mathrm{simp}}$, $L_{\mathrm{simp}}$, FWHM proportionality to $\sigma$), the ratio between the width of an anisotropic feature $x_{f,\mathrm{anis}}$ and the optimal scale $\sigma^{\ast}$ is independent of the feature size.
\label{prop:ring-line_ratio}
\end{prop}

To prove the proposition we start by finding the intersection area between the line and the ring. It can be defined using an equation for a circular segment area (Fig. \ref{fig:prop1_segment}) \cite{harris1998a}:
\begin{align}
A(x,r) = r^2 \arccos\!\left(\frac{x}{r}\right) - x\sqrt{r^2-x^2}.
\end{align}
Then this intersection is (from Fig. \ref{fig:prop1_ring_line}):
\begin{align}
A_{rl} = & A(x_1,r_2)-A(x_2,r_2)-A(x_1,r_1)+A(x_2,r_1).
\end{align}
We want to find the maximum of $A_{rl}$ in relation to $r_c$. The general derivative over r for a single A is equal to:
\begin{align}
\frac{dA}{dr} = \frac{x}{\sqrt{1-(x^2/r^2)}} + 2r \arccos{\Big(\frac{x}{r}\Big)} - \frac{rx}{\sqrt{r^2-x^2}},
\end{align}
and for positive $r$ it can be simplified to:
\begin{align}
\frac{dA}{dr} = 2r \arccos{\Big(\frac{x}{r}\Big)}.
\end{align}
As a result, the maximum of $A_{rl}$ in relation to $r_c$ can be found by setting the gradients to zero:
\begin{align}
\frac{dA_{rl}}{dr_c} = 0 
\end{align}
\begin{align}
& 2r_2 \arccos{\Big(\frac{x_1}{r_2}\Big)} - 2r_2 \arccos{\Big(\frac{x_2}{r_2}\Big)} - \notag  \\ 
& 2r_1 \arccos{\Big(\frac{x_1}{r_1}\Big)} + 2r_1 \arccos{\Big(\frac{x_2}{r_1}\Big)}  = 0.
\end{align}
Expanding the terms we have:
\begin{align}
& (\sigma a_r + \sigma w_r)\arccos{\Big(\frac{x_c - \sigma w_x}{\sigma a_r + \sigma w_r}\Big)} - \notag \\ 
&  (\sigma a_r + \sigma w_r) \arccos{\Big(\frac{x_c + \sigma w_x}{\sigma a_r + \sigma w_r}\Big)} - \notag \\ 
& (\sigma a_r - \sigma w_r) \arccos{\Big(\frac{x_c - \sigma w_x}{\sigma a_r - \sigma w_r}\Big)} + \notag \\ 
& (\sigma a_r - \sigma w_r) \arccos{\Big(\frac{x_c + \sigma w_x}{\sigma a_r - \sigma w_r}\Big)}  = 0
\end{align}

Since we are interested in extracting a ratio between the ring radius and line distance, we define $\psi = \frac{x_c}{\sigma}$, resulting in:
\begin{align}
& (a_r + w_r)\arccos{\Big(\frac{\psi-  w_x}{ a_r +  w_r}\Big)} - \notag \\ 
&  (a_r + w_r) \arccos{\Big(\frac{\psi + w_x}{ a_r + w_r}\Big)} - \notag \\ 
& ( a_r - w_r) \arccos{\Big(\frac{\psi - w_x}{a_r - w_r}\Big)} + \notag \\ 
& ( a_r - w_r) \arccos{\Big(\frac{\psi + w_x}{a_r - w_r}\Big)}  = 0
\end{align}
We can see that the value of $\psi$ depends only on constants. It can also be shown that the function $\frac{dA_{rl}}{dr_c}$ is negative for $\psi = 0$ and positive for $\psi = a_r + w_r + x_c$.

Given that $\frac{dA_{rl}}{dr_c}$ can be positive and negative, and that it is continuous in its domain, there exists a solution to $\psi$ that sets the gradients to zero. Given that this solution depends only on constants, it is also constant (independent of scale $\sigma$ or feature size $x_f$).

\end{appendices}

\section*{Acknowledgment}
Thanks to Hans Martin Kjer for sharing his insights, which led to fruitful discussions on the development of structure tensor scale-space. 

\printbibliography

\vspace{1cm}

\begin{IEEEbiographynophoto}{Pawel T. Pieta} received a B.Eng. degree in Automatics and Robotics from the AGH University of Science and Technology, Poland in 2020 and the M.Sc. degree in Autonomous Systems from the Technical University of Denmark (DTU) in 2022. He is currently pursuing a PhD degree in computer science with DTU under the supervision of Assoc. Prof. Anders N. Christensen and Prof. Anders B. Dahl. His research interests include 3D image processing, deep learning, and X-ray-based image acquisition.
\end{IEEEbiographynophoto}

\begin{IEEEbiographynophoto}{Anders B. Dahl} is a professor in 3D image analysis, and head of the Section for Visual Computing, Department of Applied Mathematics and Computer Science, Technical University of Denmark (DTU). He is heading the Center for Quantification of Imaging Data (QIM) from MAX IV, focusing on quantitative analysis of 3D images. His research is focused on image segmentation and its applications.
\end{IEEEbiographynophoto}

\begin{IEEEbiographynophoto}{Jeppe Revall Frisvad} is an associate professor with more than 15 years of experience in material appearance modeling and rendering. As a highlight, his work includes the first directional dipole model for subsurface scattering, and his research includes methods for both computation and photographic measurement of the optical properties of materials.
\end{IEEEbiographynophoto}

\begin{IEEEbiographynophoto}{Siavash A. Bigdeli} 
is an associate professor of computer vision at the Technical University of Denmark (DTU). Prior to that, he was a scientist at CSEM and a postdoc at EPFL.
He works in the area of machine learning to build explicit and explainable statistical models for Computer Vision and Graphics applications.
His interests lie in the Ante-/Post-Hoc explainability of ML models by employing statistical models in learning/inference processes, and by integrating philosophical methods in ML with a focus on Generative AI.
\end{IEEEbiographynophoto}

\begin{IEEEbiographynophoto}{Anders N. Christensen} is an associate professor in image analysis and multivariate statistics at the Section for Visual Computing, Department of Applied Mathematics and Computer Science, Technical University of Denmark (DTU). His research is focused on explainable image analysis and its applications in both the medical and material science domains.
\end{IEEEbiographynophoto}

\vfill

\EOD

\end{document}